\documentclass[10pt,letterpaper,logo,onecolumn,teaser]{CAFE/cafestyle}
\usepackage[T1]{fontenc}
\usepackage{microtype}
\usepackage{nicefrac}
\usepackage{xspace}
\usepackage{amsmath, amssymb, amsfonts, amsthm, mathtools, mathrsfs}
\usepackage{dsfont}
\usepackage{fdsymbol}

\usepackage{booktabs}
\usepackage{nicematrix}
\usepackage{multirow}
\usepackage{makecell}
\usepackage{bigstrut}
\usepackage{enumitem}
\setitemize{label=\textbullet, leftmargin=*, nolistsep}
\usepackage{graphicx}
\usepackage{adjustbox}
\usepackage{float}
\usepackage{wrapfig}
\usepackage{subcaption}
\expandafter\def\csname ver@subfig.sty\endcsname{}
\usepackage[dvipsnames]{xcolor}
\usepackage{color}
\usepackage{fontawesome5}
\usepackage{pifont}
\usepackage{hyperref}
\hypersetup{colorlinks=true, citecolor=HKUTeal}
\usepackage[all]{hypcap}
\usepackage{cleveref}
\usepackage{url}
\usepackage{algorithm}
\usepackage{algorithmicx}
\usepackage{algpseudocode}
\usepackage[normalem]{ulem}
\usepackage{lipsum}
\usepackage{rotating}
\usepackage{stackengine}
\usepackage{caption}
\usepackage{listings}
\usepackage{soul}
\usepackage{gradient-text}
\usepackage{pgfplots}
\pgfplotsset{compat=newest}
\usepackage{pgfplotstable}
\usepackage{tikz}
\usepackage{svg}

\usepackage[comma,numbers,sort,compress]{natbib}

\hypersetup{
    colorlinks = true,
    citecolor  = {HKUTeal},
    linkcolor  = {HKUGreen},
    urlcolor   = {HKUGreen},
}
\definecolor{demphcolor}{RGB}{125,125,125}             

\tcbuselibrary{breakable, skins}
\newtcolorbox{hkubox}[1]{
  enhanced,
  breakable,
  colback=white,
  colframe=HKUGreen!85,
  coltitle=HKUGreen,
  fonttitle=\bfseries\sffamily,
  title=#1,
  titlerule=0.8pt,
  boxrule=1pt,
  left=3mm, right=3mm, top=2mm, bottom=2mm,
  boxsep=1mm,
  before upper=\smallskip,
}

\crefname{equation}{Eq.}{Eqs.}
\crefformat{section}{\S#2#1#3}
\crefformat{subsection}{\S#2#1#3}
\crefformat{subsubsection}{\S#2#1#3}
\crefrangeformat{section}{\S\S#3#1#4 to~#5#2#6}
\crefmultiformat{section}{\S\S#2#1#3}{ and~#2#1#3}{, #2#1#3}{ and~#2#1#3}

\usepackage{fancyhdr}
\PassOptionsToPackage{pagebackref,breaklinks,colorlinks,allcolors=HKUGreen}{hyperref}

\usepackage[T1]{fontenc}
\usepackage{microtype}
\usepackage{nicefrac}
\usepackage{xspace}

\usepackage{amsmath, amssymb, amsfonts, amsthm, mathtools, mathrsfs}
\usepackage{dsfont}
\usepackage{fdsymbol}

\usepackage{booktabs}
\usepackage{nicematrix}
\usepackage{multirow}
\usepackage{makecell}
\usepackage{bigstrut}
\usepackage{enumitem}
\setitemize{label=\textbullet, leftmargin=*, nolistsep}
\usepackage{diagbox}
\usepackage{seqsplit}

\usepackage{graphicx}
\usepackage{adjustbox}
\usepackage{float}
\usepackage{wrapfig}
\usepackage{subcaption}
\expandafter\def\csname ver@subfig.sty\endcsname{}

\usepackage{color}
\usepackage{fontawesome5}
\usepackage{pifont}

\usepackage{hyperref}
\hypersetup{colorlinks=true, citecolor=HKUTeal}
\usepackage[all]{hypcap}
\usepackage{url}

\usepackage{algorithm}
\usepackage{algorithmicx}
\usepackage{algpseudocode}

\usepackage[normalem]{ulem}
\usepackage{lipsum}
\usepackage{rotating}
\usepackage{stackengine}
\usepackage{caption}
\usepackage{listings}
\usepackage{soul}
\usepackage{gradient-text}
\usepackage{textcomp}
\usepackage{fancyvrb}

\DefineVerbatimEnvironment{toolcall}{Verbatim}
  {breaklines=true, breakanywhere=true,
   fontsize=\small, frame=none, xleftmargin=0pt}

\usepackage{pgfplots}
\pgfplotsset{compat=newest}
\usepackage{pgfplotstable}
\usepackage{tikz}
\usepackage{svg}

\usepackage[most]{tcolorbox}
\tcbuselibrary{breakable, skins}

\definecolor{cafemim}{HTML}{B58A36}  
\definecolor{cafeont}{HTML}{1F5F8B}  
\definecolor{cafectx}{HTML}{2E8B57}  

\tcbset{
  cafeSubtitle/.style={
    enhanced, breakable,
    colback=cafemim!4,
    colframe=cafemim!50!black,
    boxrule=0pt,
    left=2pt,right=2pt,top=2pt,bottom=2pt,
  },
  cafeSubTitleLine/.style={
    colback=cafemim!70!black,
    colframe=cafemim!70!black,
    coltitle=white,
    fonttitle=\bfseries,
    boxrule=0pt,
    left=6pt,right=6pt,top=3pt,bottom=3pt,
  }
}

\usepackage{bbding}

\title{%
  \raisebox{-0.6ex}{\includegraphics[height=3.8ex]{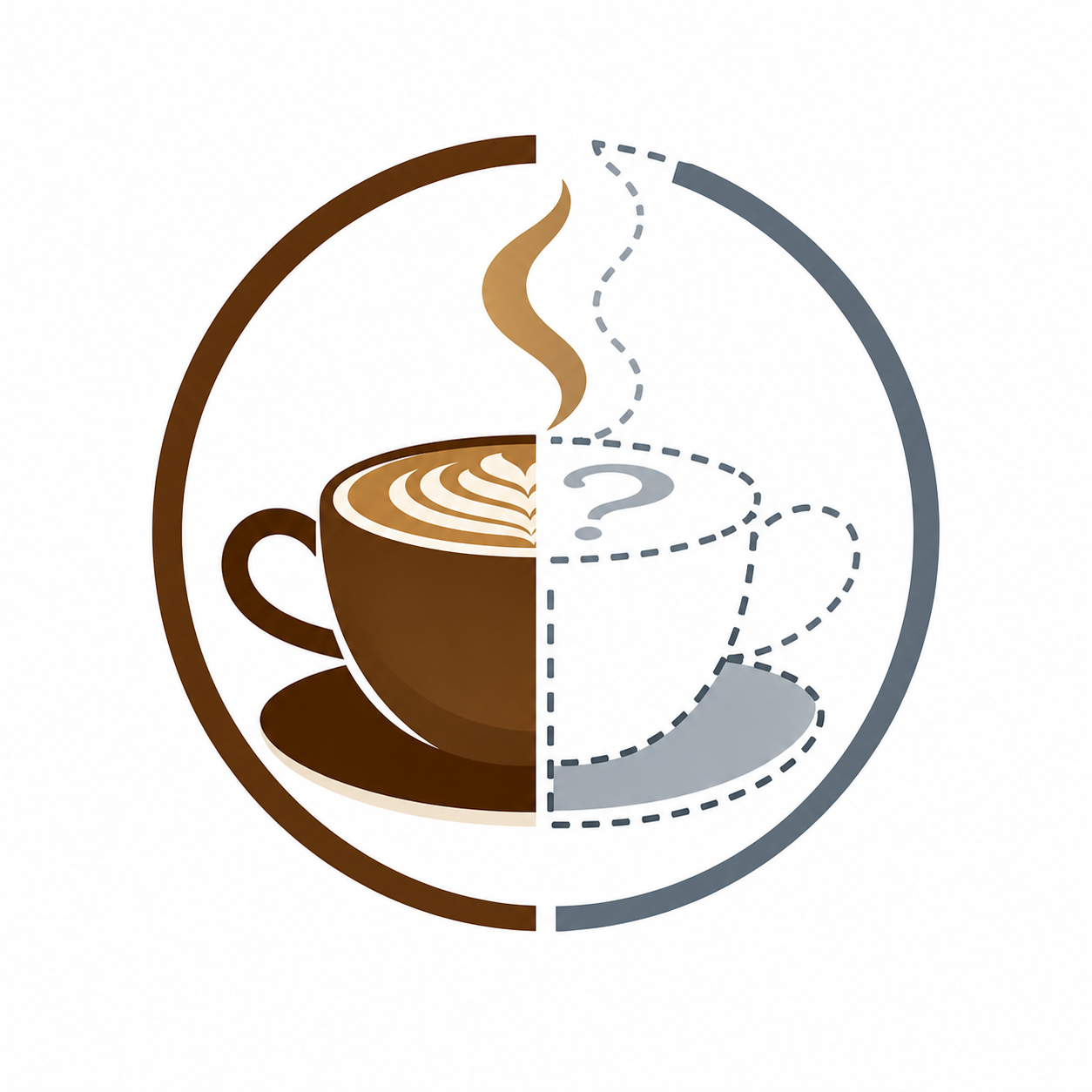}}\hspace{0.2em}%
  From Pixels to Concepts: Do Segmentation Models Understand What They Segment?
}

\author{
\begin{tabular}{c}
\textbf{Shuang Liang$^{1,3,\dagger}$, Zeqing Wang$^{2,\dagger}$, Yuxian Li$^{1,\dagger}$, Xihui Liu$^{1}$, Han Wang$^{1,3\ast}$}
\end{tabular}
}

\affil{
  $^{1}$Department of Electrical and Computer Engineering, The University of Hong Kong \\
$^{2}$School of Computer Science and Engineering, Sun Yat-sen University \\
  $^{3}$CASIC, The University of Hong Kong
}

\correspondingauthor{
\begin{tabular}{@{}l@{}}
\small $^{\dagger}$ Equal contribution. \\
\small $^{\ast}$ Corresponding author.
\end{tabular}
}

\renewcommand{\insertteaserfigure}{
\begin{center}
    \centering
    \captionsetup{type=figure}
    \includegraphics[width=1.0\linewidth]{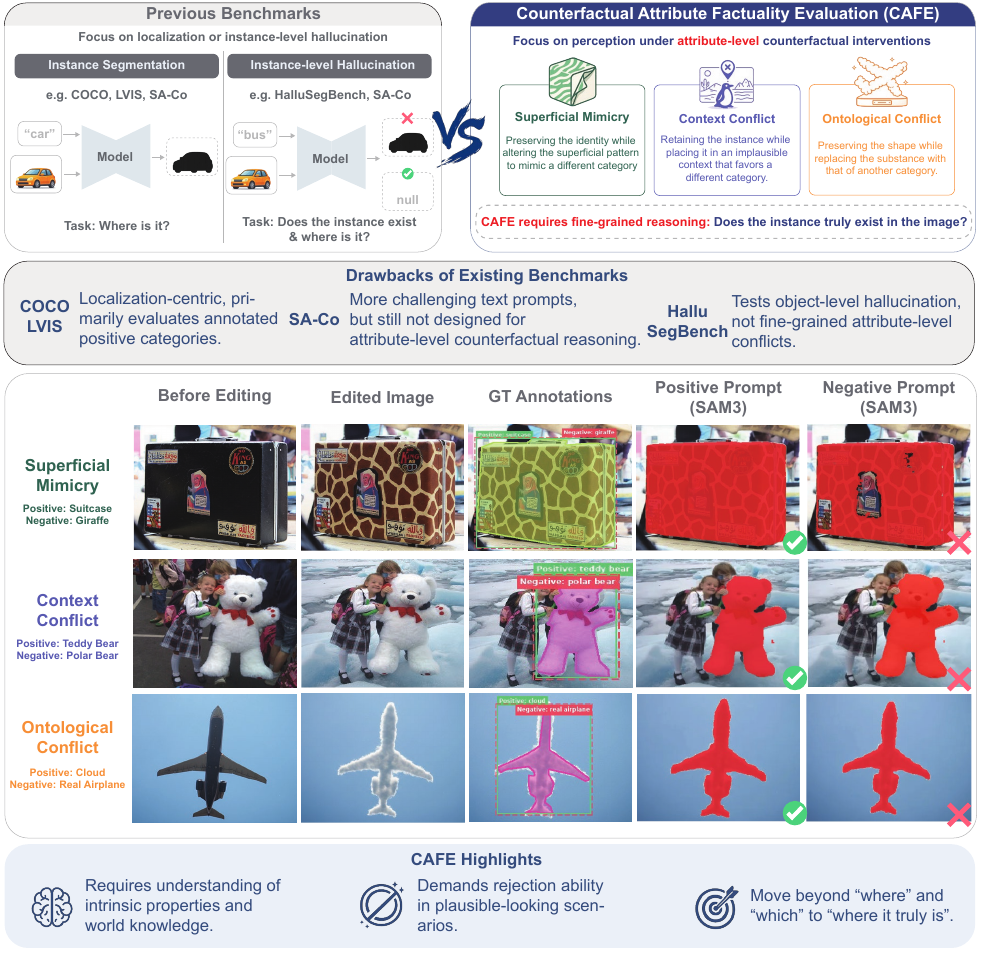}
    \caption{
    \textbf{CAFE focuses on attribute-level counterfactual evaluation,} where specific visual attributes are modified within a fixed target mask inherited from the original annotation. \textit{Positive prompts} refer to the semantically valid concept for the edited target region, while \textit{negative prompts} refer to misleading but semantically invalid concepts. The original object remains the positive concept in \textit{Superficial Mimicry} and \textit{Context Conflict}; for \textit{Ontological Conflict} the converted material becomes the positive concept and the original object becomes the misleading negative concept. For example, after replacing the background of a teddy bear with a snowy scene, SAM3 correctly segments the object when prompted with the positive prompt ``teddy bear''. However, when given the negative prompt ``polar bear'', which is plausible under the surrounding visual context but semantically inconsistent with the object identity, SAM3 still segments the same instance with high confidence.}
    \label{fig:cafe_overview}
    \vspace{-0.2cm}
\end{center}
}

\begin{document}

\begin{abstract}

Segmentation is a fundamental vision task underlying numerous downstream applications. Recent promptable segmentation models, such as Segment Anything Model 3 (SAM3), extend segmentation from category-agnostic mask prediction to concept-guided localization conditioned on high-level textual prompts. However, existing benchmarks primarily evaluate mask accuracy or object presence, leaving unclear whether these models faithfully ground the queried concept or instead rely on visually salient but semantically misleading cues. We introduce CAFE: \textbf{C}ounterfactual \textbf{A}ttribute \textbf{F}actuality \textbf{E}valuation, a novel benchmark for evaluating concept-faithful segmentation in promptable segmentation models. Our \textbf{CAFE} is built on attribute-level counterfactual manipulation: the target region and ground-truth mask are preserved, while attributes such as surface appearance, context, or material composition are modified to introduce misleading semantic cues. The benchmark contains 2,146 paired test samples, each consisting of a target image, a ground-truth mask, a positive prompt, and a misleading negative prompt. These samples cover three counterfactual categories: Superficial Mimicry (\textbf{SM}), Context Conflict (\textbf{CC}), and Ontological Conflict (\textbf{OC}). We evaluate various model types and sizes on our CAFE. Experiments reveal a systematic gap between localization quality and concept discrimination: models often generate accurate masks even for misleading prompts, suggesting that strong mask prediction does not necessarily imply faithful semantic grounding. Our CAFE provides a controlled benchmark for diagnosing whether promptable segmentation models perform concept-faithful grounding rather than shortcut-driven mask retrieval.

\textbf{Project Page:} \url{https://t-s-liang.github.io/CAFE} \\
\textbf{Code:} \url{https://github.com/T-S-Liang/CAFE} \\
\textbf{Dataset:} \url{https://huggingface.co/datasets/teemosliang/CAFE}

\end{abstract}

\maketitle

\section{Introduction}

Segmentation has long been a central problem in computer vision, evolving from category-level dense prediction in semantic segmentation~\cite{chen2017deeplab, xie2021segformer}, to instance-aware mask prediction~\cite{he2017mask, chen2018masklab, michaelis2018one}, and more recently to open-vocabulary and promptable segmentation~\cite{ghiasi2022scaling, zhang2023simple, ren2024grounded, xu2023open}. This progression relaxes closed-set categories and enables a prompt-guided region association.

Early promptable segmentation models, such as SAM~\cite{kirillov2023segment} and SAM2~\cite{ravi2024sam}, focus on visual prompts, such as points, boxes and primarily address spatial grounding without explicit textual concept conditioning. In parallel, open-vocabulary segmentation and grounding-segmentation pipelines use language queries to localize semantic regions, often by coupling a grounding or detection model, such as Grounding DINO~\cite{liu2024grounding}, with a mask generator~\cite{ren2024grounded}. Recently, SAM3~\cite{carion2025sam} introduced promptable concept segmentation (PCS), an end-to-end formulation that directly produces masks from concept prompts, without relying on an explicit grounding or detection stage to generate intermediate boxes.

Standard benchmarks such as COCO~\cite{lin2014microsoft}, ADE20K~\cite{zhou2019semantic}, and LVIS~\cite{gupta2019lvis} primarily evaluate segmentation accuracy over predefined visual categories. Recent counterfactual benchmarks, such as HalluSegBench~\cite{li2025counterfactual} further tests object-level counterfactual hallucination by pairing factual images with counterfactual images in which the referred object is absent.
However, counterfactual segmentation is not limited to object-level presence or absence. Fine-grained conflicts can arise when the target region remains visible and localizable, but attributes that affect concept identity, such as surface appearance, surrounding context, or material composition, are deliberately modified. In this setting, a model may produce a geometrically accurate mask for a semantically invalid prompt. Existing benchmarks therefore provide limited diagnosis of whether promptable segmentation models distinguish concept-faithful grounding from shortcut-driven responses to misleading attribute cues.

\begin{figure}[!t]
  \centering
 
  \includegraphics[width=0.9\textwidth]{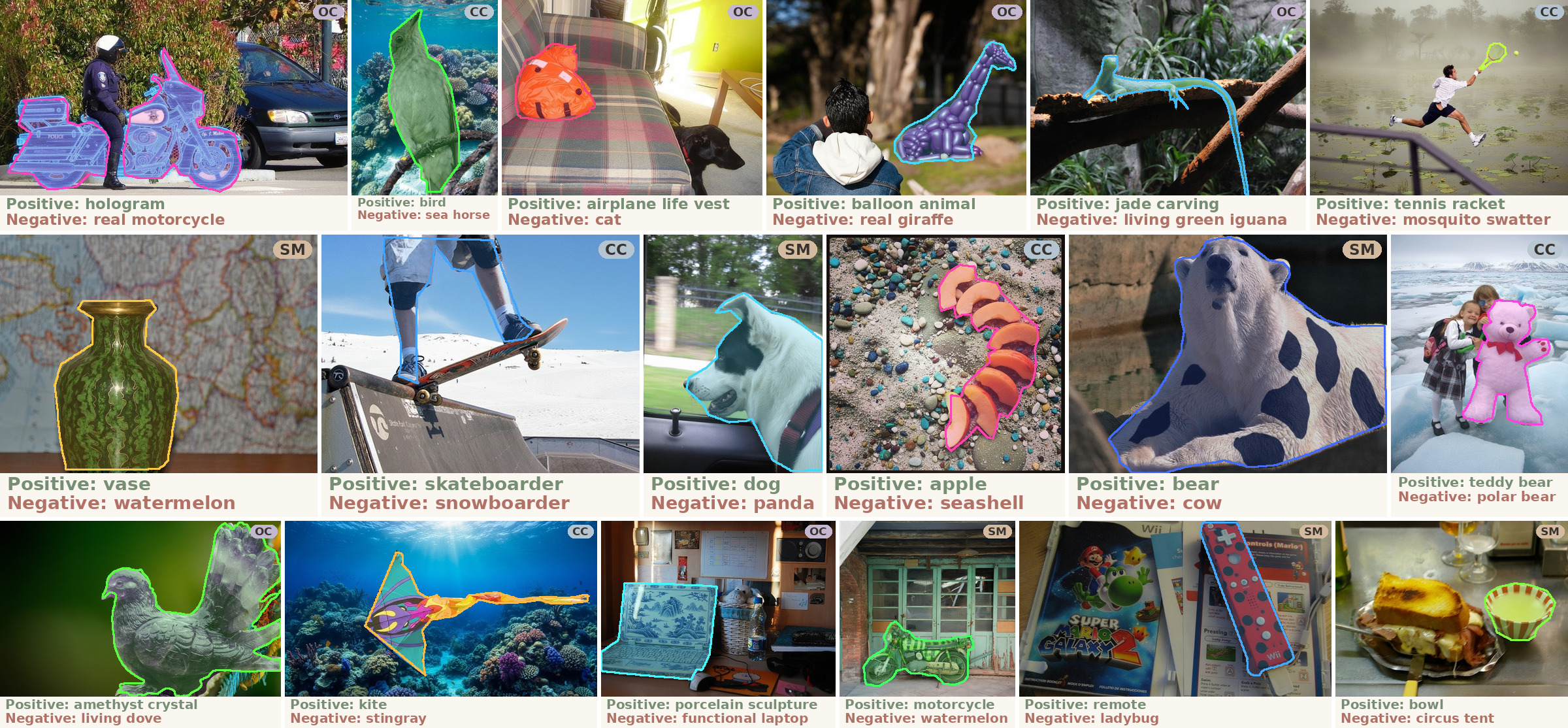}
  \caption{Examples in our CAFE. Each sample contains a counterfactually edited target image, a ground-truth mask for the target region, a positive prompt that is semantically valid for the target, and a misleading negative prompt that is visually plausible but semantically invalid. The examples cover three attribute-level intervention types: Superficial Mimicry (\textbf{SM}), Ontological Conflict (\textbf{OC}), and Context Conflict (\textbf{CC}).}
  \label{fig:exp_anns}
\end{figure}

To this end, we propose \textbf{CAFE}, the \textbf{C}ounterfactual \textbf{A}ttribute \textbf{F}actuality \textbf{E}valuation for promptable segmentation models. CAFE preserves the target region and its annotation mask while counterfactually manipulating attributes that affect concept identity, including surface appearance, surrounding context, and material composition. This design tests whether model responses remain consistent with human semantic judgments when the target region remains localizable but contains misleading attribute cues.
We design three categories of attribute-level interventions: \textit{superficial mimicry}, \textit{context conflict}, and \textit{ontological conflict}. Each intervention preserves the target region and its segmentation mask while modifying one attribute dimension that affects concept identity. \textit{Superficial mimicry} modifies surface appearance to make the target visually resemble another category while preserving its underlying object identity. \textit{Context conflict} modifies the surrounding context to introduce environmental evidence associated with another category while preserving the target object's identity. \textit{Ontological conflict} modifies material composition so that the target region changes its substance while preserving its global shape. These interventions create cases where the target remains localizable, but the misleading negative prompt is semantically invalid according to human judgment despite being supported by salient attribute cues.
Fig.~\ref{fig:cafe_overview} shows representative examples. These examples demonstrate that promptable segmentation models may produce confident masks for semantically invalid negative prompts when the edited target remains localizable and contains misleading attribute cues. In \textit{superficial mimicry}, a suitcase is painted with giraffe-like patterns while its object identity remains a suitcase. The positive prompt is therefore ``suitcase'', whereas the misleading negative prompt is ``giraffe'', which is supported only by the edited surface appearance.
In \textit{context conflict}, a teddy bear is placed in a snowy scene while its object identity remains a teddy bear. The positive prompt remains ``teddy bear'', whereas the misleading negative prompt is ``polar bear'', which is supported by the edited surrounding context rather than the target object itself.
In \textit{ontological conflict}, an airplane-shaped target is re-rendered as cloud while preserving its global shape. The target region is therefore materially a cloud rather than an airplane. In this case, the positive prompt is ``cloud'', whereas the misleading negative prompt is ``real airplane'', which is supported only by the retained global shape rather than the material composition of the edited target.
We collect source images and annotations from COCO~\cite{lin2014microsoft}, LVIS~\cite{gupta2019lvis}, and SA-Co/Gold~\cite{carion2025sam}, and perform controlled attribute-level image editing using category-specific prompts. After multi-stage filtering and validation by three human annotators, CAFE contains 2,146 paired test samples. Each test sample consists of a target image, a ground-truth mask, a positive prompt that describes a semantically valid concept, and a misleading negative prompt that is visually plausible but semantically invalid for the target region.

Our contributions are summarized as follows: i) We introduce \textbf{CAFE}, a benchmark for evaluating concept-faithful grounding in promptable segmentation models under controlled counterfactual attribute interventions. CAFE covers three categories of attribute-level semantic conflict, namely \textit{superficial mimicry}, \textit{context conflict}, and \textit{ontological conflict}, which, respectively, manipulate surface appearance, surrounding context, and material composition while preserving the target region and its annotation mask. ii) We construct 2,146 paired test cases, each containing an edited target image, a ground-truth mask, a positive prompt, and a misleading negative prompt. All cases are validated by human annotators to ensure that the target region remains localizable and that the positive and negative prompts reflect clear human semantic judgments under the edited attributes. iii) We evaluate both end-to-end promptable concept segmentation models, such as SAM3, framework-based open-vocabulary grounding-segmentation pipelines, such as Grounded SAM2, and an agentic verification variant that uses SAM3 as a segmentation tool, denoted as CAFE-SAM3. The results reveal a systematic gap between mask localization quality and concept-faithful grounding: current models can produce accurate masks for misleading negative prompts, indicating that they often respond to salient attribute cues rather than the semantic validity of the queried concept.

\section{Related Works}

\textbf{Counterfactual Evaluation for Pixel-Level Grounding.}  Counterfactual evaluation has been widely used to assess whether model predictions rely on causal evidence rather than spurious correlations. Prior work has applied counterfactual or minimally edited inputs to evaluate fairness, robustness, and vision-language understanding~\cite{kusner2017counterfactual,kaushik2019learning,li2024naturalbench, wang2025generated, wang2025phydetex, wang2025timecausality}. Recent work has begun to examine this issue in segmentation. Generalized referring expression segmentation extends the classical single-target setting to no-target and multi-target expressions, requiring models to decide whether a queried concept is visually grounded before producing a mask~\cite{liu2023gres}. Counterfactual segmentation benchmarks further diagnose pixel-grounding hallucinations by constructing factual and counterfactual pairs, where models should segment the target in the factual image but abstain when the target object is removed or replaced~\cite{li2025counterfactual}. Our CAFE follows this counterfactual perspective but studies a finer-grained and complementary setting: the target region remains visible and localizable, while attributes such as appearance, material, or context are manipulated. This design tests whether such models faithfully ground the queried concept rather than relying on misleading attribute cues.

\textbf{Open-Vocabulary and Promptable Segmentation.} Classical semantic and instance segmentation models are typically trained and evaluated under a closed-vocabulary setting, where categories are predefined. SAM~\cite{Kirillov_2023_ICCV} and SAM2~\cite{ravi2024sam} relax this paradigm by formulating segmentation as class-agnostic promptable mask prediction, where users provide visual prompts. SAM2 further extends this formulation to video through a memory-based promptable segmentation architecture. A parallel line of work introduces language into segmentation by combining open-vocabulary detectors or grounding models, such as Grounding DINO~\cite{liu2024grounding} and OWLv2~\cite{minderer2023scaling}. More recent methods move toward unified open-vocabulary segmentation. YOLO-World~\cite{cheng2024yolo} improves open-vocabulary detection through vision-language modeling and large-scale region-text pretraining, and extends to instance segmentation with an additional segmentation head. OpenSeeD~\cite{zhang2023simple} jointly learns detection and segmentation in a shared semantic space. SAM3~\cite{carion2025sam} further formulates promptable concept segmentation, directly producing masks from concept prompts such as noun phrases, image exemplars, or their combinations. These advances make it increasingly important to evaluate not only whether models can produce accurate masks, but also whether their masks are semantically faithful to the input prompt.

\textbf{Benchmarking Segmentation Models.} Segmentation benchmarks have evolved along two axes: output granularity from semantic~\cite{long2015fully} to instance~\cite{he2017mask,hariharan2014simultaneous} and panoptic segmentation~\cite{kirillov2019panoptic} and interaction paradigm—from closed-vocabulary~\cite{he2017mask} to visual promptable~\cite{kirillov2023segment,ravi2024sam}, language-guided or open-vocabulary~\cite{ren2024grounded,openworldsam2025}, and promptable concept segmentation~\cite{carion2025sam}. Most benchmarks, such as COCO~\cite{lin2014microsoft} and LVIS~\cite{gupta2019lvis}, focus on mask overlap metrics (IoU, AP, AR), which only measure spatial accuracy. Some other benchmarks like RefCOCO and RefCOCOg~\cite{mao2016generation,yu2016modeling,kazemzadeh2014referitgame} evaluate language-guided localization but do not test whether models reject semantically unsupported or counterfactual queries. SA-Co~\cite{carion2025sam} and HalluSegBench~\cite{li2025hallusegbench} partially address semantic grounding, with HalluSegBench using factual and counterfactual object replacement to reveal pixel-grounding hallucinations. Our CAFE complements these benchmarks by evaluating \textbf{attribute-level} semantic validity under \textbf{mask-preserving} counterfactual edits: the target region remains visible and annotated while appearance or material is manipulated, exposing cases where models produce accurate masks for misleading prompts and revealing shortcut-driven mask retrieval rather than concept-faithful grounding.

\section{Task Definition}

In this section, we formalize the task of evaluating counterfactual attribute factuality for segmentation models. In this work, a \textit{counterfactual} image is defined as an edited version of an original image in which a specific attribute of the target region is deliberately changed from its factual state to an alternative state, while the target region remains spatially identifiable and serves as the evaluation anchor. The semantically valid concept after editing may either preserve the original object identity or shift to a new material- or substance-defined concept, depending on the type of counterfactual manipulation. This controlled edit introduces a visually plausible but semantically invalid competing concept, enabling us to evaluate whether a segmentation model follows the semantically valid concept in the edited image or incorrectly responds to the counterfactually induced cue.
We define three categories of counterfactual scenarios in which specific visual attributes are manipulated, including superficial patterns, surrounding visual contexts, and substances or materials.

\subsection{Counterfactual Attribute Scenarios}
\textbf{Superficial Mimicry.} The superficial pattern of an object is repainted or covered with a confusing pattern associated with another kind of object. For example, as shown in Fig.~\ref{fig:exp_anns}, the \textit{vase} is recolored with the pattern of a \textit{watermelon}, thereby creating a misleading counterfactual cue while keeping the concept of vase semantically valid. The positive prompt therefore refers to the object itself, whereas the misleading negative prompt refers to the repainted superficial pattern.

\textbf{Context Conflict.} The visual surroundings of an object are replaced with another environment that is implausible for the object. For example, as shown in Fig.~\ref{fig:exp_anns}, the skateboarder is placed in a snowy environment. The positive prompt remains \textit{skateboarder}, while the misleading prompt is \textit{snowboarder}, since the person appearing in this scenario is highly plausible as a snowboarder. More generally, in context-conflict cases, the positive prompt refers to the original object identity, while the misleading negative prompt refers to a contextually plausible but semantically invalid concept suggested by the swapped environment.

\textbf{Ontological Conflict.} The substance of the original object is re-rendered and replaced by another kind of material. For example, as shown in Fig.~\ref{fig:exp_anns}, the living dove is re-rendered as a crystal sculpture. The positive prompt is therefore \textit{amethyst crystal}, while the misleading negative prompt is \textit{living dove}. In general, the positive prompt refers to the re-rendered material or substance, whereas the misleading negative prompt refers to the original object identity that is no longer semantically valid.

\subsection{Prompt Pair Construction}
For each counterfactual scenario, we construct a pair of prompts: a positive prompt $q^{+}$ and a misleading negative prompt $q^{-}$. The positive prompt refers to the semantically valid concept in the edited image, while the misleading negative prompt refers to a visually plausible but semantically invalid concept induced by counterfactual manipulation. Therefore, each sample is represented as a tuple $(I, M^{\star}, q^{+}, q^{-}, c)$, where $I$ denotes the edited image, $M^{\star}$ denotes the target mask, $q^{+}$ and $q^{-}$ denote positive and misleading negative prompts, and $c$ denotes the counterfactual category.

\subsection{Semantic Validity}
We define semantic validity as whether the queried concept is supported by visual evidence in the edited image. For each sample, the positive prompt is semantically valid, while the misleading negative prompt is semantically invalid. Formally, let $v(I,q) \in \{0,1\}$ indicate whether the query $q$ is semantically valid in the image $I$. By construction, each sample satisfies
\begin{equation}
v(I,q^{+}) = 1, \qquad v(I,q^{-}) = 0.
\end{equation}

\subsection{Evaluation Objective}
\label{sec:eval_objective}

Given a segmentation model $f$, an image $I$, and a query $q$, the model produces a predicted mask $\hat{M}=f(I,q)$ with a confidence score $s$. The goal is to evaluate whether the model can localize the target under the positive prompt while rejecting the misleading concept under the negative prompt. Under the positive prompt $q^{+}$, the model is expected to produce a high-confidence target-aligned prediction,
\begin{equation}
\operatorname{IoU}(f(I,q^{+}), M^{\star}) \geq \tau
\quad \text{and} \quad
s(f(I,q^{+})) \geq t.
\end{equation}
Under the misleading negative prompt $q^{-}$, the model is expected to reject the query by assigning a confidence score below the acceptance threshold,
\begin{equation}
s(f(I,q^{-})) < t.
\end{equation}
If the model instead produces a high-confidence prediction under $q^{-}$, we further use its overlap with the target mask $M^{\star}$ to distinguish whether the false positive is target-aligned or unaligned. Here, $\tau$ denotes the IoU threshold used to determine target alignment, and $t$ denotes the confidence threshold used to determine whether a prediction is accepted as a positive response. The full classification protocol is formalized in Section~\ref{sec:metrics}.

\begin{figure}[!t]
  \centering
  \includegraphics[width=1.0\textwidth]{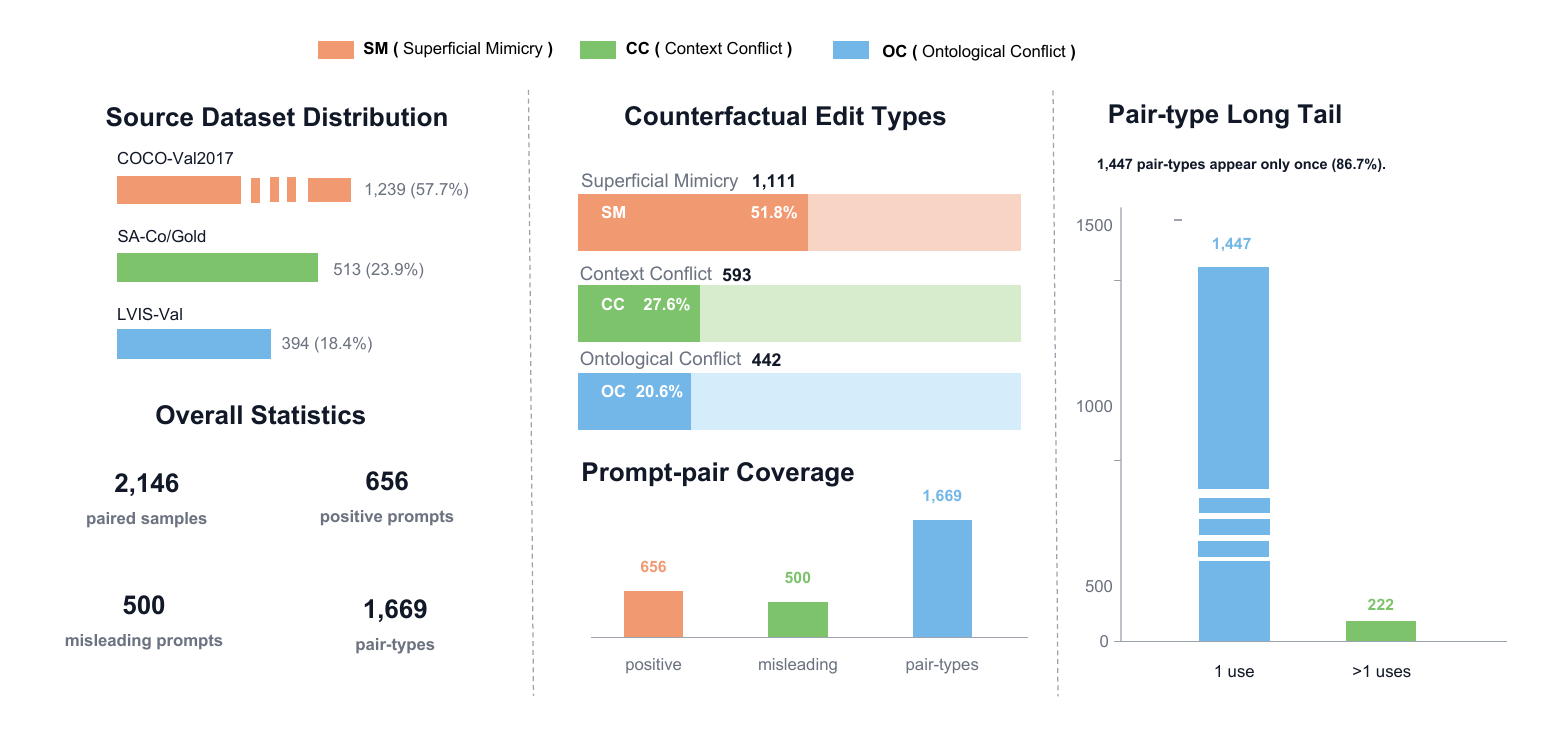}
  \caption{Overview of CAFE benchmark statistics. CAFE contains 2,146 paired counterfactual samples from three source datasets and spans three edit types: superficial mimicry (SM), context conflict (CC), and ontological conflict (OC). 
CAFE provides 656 positive prompts and 500 misleading prompts, forming 1,669 prompt pairs whose distribution is highly long-tailed, with 1,447 pair types appearing only once, indicating broad semantic coverage across counterfactual concept pairs.
}
  \label{fig:statistics}
\end{figure}

\section{CAFE: \textbf{C}ounterfactual \textbf{A}ttribute \textbf{F}actuality \textbf{E}valuation}

\subsection{Dataset Statistics}

Fig.~\ref{fig:statistics} summarizes CAFE, which contains 2,146 paired counterfactual samples drawn from COCO-Val2017~\cite{lin2014microsoft} (1,239 samples), SA-Co/Gold~\cite{carion2025sam} (513), and LVIS-Val~\cite{gupta2019lvis} (394), combining common object categories with diverse open-vocabulary concepts. CAFE covers three counterfactual edit types: Superficial Mimicry (SM, 1,111 samples), where target appearance is altered with misleading surface patterns; Context Conflict (CC, 593), where target placement or surroundings suggest a misleading context; and Ontological Conflict (OC, 442), where visual evidence implies a semantically incompatible category or material. These edits test whether segmentation models can reject prompts that are visually plausible but semantically invalid. CAFE includes 656 positive prompts and 500 misleading prompts, forming 1,669 prompt pairs. The pair-type distribution is long-tailed: 1,447 pairs (86.7\%) appear only once, limiting over-reliance on frequent concept pairs and providing broad coverage of counterfactual semantic relations. Details of the annotation pipeline are in Appendix~\ref{appendix:dataset_gen}.

\subsection{Evaluation Metrics}
\label{sec:metrics}

\begin{table}[t]
\centering
\small
\caption{
Target-aware classification used in CAFE. Each annotation is paired with a positive prompt ($p^{+}$) and a misleading negative prompt ($p^{-}$). A prediction is accepted as a positive response when its confidence score satisfies $s\geq t$, where $t$ denotes the presence confidence threshold. A prediction is target-aligned when its overlap with the target mask satisfies $\operatorname{IoU}\geq\tau$, where $\tau$ denotes the IoU threshold. TA denotes target-aligned cases, and UA denotes unaligned cases. Under $p^{+}$, TA-TP denotes a successful target-aligned positive prediction, while UA-P denotes a high-confidence but unaligned positive response. Although UA-P indicates that the model responds to the positive prompt, it fails to localize the target and is therefore counted as a false negative for target-aware evaluation. Under $p^{-}$, high-confidence responses are false positives, further separated into TA-FP and UA-FP according to their target alignment. Low-confidence responses are counted as TN because the misleading prompt is rejected.
}
\label{tab:cafe-binclass}
\setlength{\tabcolsep}{6pt}
\renewcommand{\arraystretch}{1.25}
\begin{tabular}{cc}
\begin{tabular}{c|c|c}
\multicolumn{3}{c}{\textbf{(a) Positive prompt } $p^{+}$}\\[2pt]
\diagbox[width=5.8em]{$s$}{$\operatorname{IoU}$} & $\,\geq \tau\,$ & $\,< \tau\,$ \\\hline
$\geq t$ & $\mathrm{TA\text{-}TP}$ & $\mathrm{UA\text{-}P}$ \\\hline
$<    t$ & $\mathrm{TA\text{-}FN}$ & $\mathrm{UA\text{-}FN}$ \\
\end{tabular}
&
\begin{tabular}{c|c|c}
\multicolumn{3}{c}{\textbf{(b) Negative prompt } $p^{-}$}\\[2pt]
\diagbox[width=5.8em]{$s$}{$\operatorname{IoU}$} & $\,\geq \tau\,$ & $\,< \tau\,$ \\\hline
$\geq t$ & $\mathrm{TA\text{-}FP}$ & $\mathrm{UA\text{-}FP}$ \\\hline
$<    t$ & $\mathrm{TN}$ & $\mathrm{TN}$ \\
\end{tabular}
\end{tabular}
\vspace{2pt}\\
\end{table}

\textbf{Class-gated F$_1$.} We follow the PCS evaluation protocol of SAM3~\cite{carion2025sam}, where cgF$_1$ combines image-level concept recognition with localization quality.
For each image-prompt pair, the model first makes a binary present/absent decision according to whether any prediction exceeds the decision threshold.
Image-level concept recognition is summarized by IL\text{-}MCC, i.e., the Matthews correlation coefficient computed over these binary concept-presence decisions.
The quality of localization is measured by positive micro F$_1$ (pmF$_1$), which evaluates mask matching in positive pairs where the queried concept is present.
cgF$_1$ combines IL\text{-}MCC and pmF$_1$ into a single calibrated operating-point score, penalizing both missing valid concepts and false acceptance of invalid prompts. For SAM3, we set the presence-confidence threshold to 0.5, following its default setting. For the remaining models, which do not include a presence head for calibration, we calibrate the threshold using a protocol similar to the SAM3 benchmark. Details of the calibration procedure are provided in Appendix~\ref{app:calibration}.

\textbf{Target-aware Classification.} We formalize the target-aware classification definitions used in CAFE. In our dataset, each ground-truth annotation is paired with a positive prompt and a carefully designed misleading negative prompt. The classification table is shown in Table~\ref{tab:cafe-binclass}. Let $\tau$ denote the IoU threshold for target alignment, and let $t$ denote the threshold for the presence confidence score $s$. Given a positive prompt, if the predicted mask aligns with the ground truth, namely if its IoU is greater than or equal to $\tau$, and the presence confidence score is greater than or equal to $t$, we count it as a target-aligned true positive (TA-TP). If the predicted mask aligns with the ground truth but the presence confidence score is lower than $t$, we count it as a target-aligned false negative (TA-FN). If the predicted mask does not align with the ground truth, namely if its IoU is lower than $\tau$, we count it as an unaligned false negative (UA-FN), regardless of whether the presence confidence score is greater than or equal to $t$. Given a misleading negative prompt, rejection is determined by the presence confidence score. A prediction with $s<t$ is counted as a true negative (TN), regardless of its IoU with the target mask $M^{\star}$. A prediction with $s\geq t$ is counted as a false positive response. We further use $\operatorname{IoU}(\hat{M},M^{\star})$ to distinguish its spatial attribution: if $\operatorname{IoU}(\hat{M},M^{\star})\geq\tau$, it is counted as a target-aligned false positive (TA-FP); otherwise, it is counted as an unaligned false positive (UA-FP).

\begin{table*}[!t]           
\centering
\caption{
Promptable Concept Segmentation (PCS) performance on CAFE.
We evaluate open-vocabulary segmentation systems under three paradigms:
end-to-end models, multi-model frameworks, and agentic methods.
We report the standard PCS metrics~\cite{carion2025sam}, including cgF$_1$, IL\_MCC, and pmF$_1$.
Results are reported for three counterfactual categories:
Superficial Mimicry (SM), Context Conflict (CC), and Ontological Conflict (OC);
\emph{Overall} aggregates over all categories.
CAFE-SAM3 (GPT-5.5) substantially improves over direct SAM~3, especially on Ontological Conflict, indicating that explicit agentic verification helps reject semantically invalid masks under misleading prompts.
}
\begin{adjustbox}{width=1\linewidth}
\begin{tabular}{l cccc cccc cccc}
\toprule
 & \multicolumn{4}{c}{cgF$_1$ $\uparrow$}
 & \multicolumn{4}{c}{IL\_MCC $\uparrow$}
 & \multicolumn{4}{c}{pmF$_1$ $\uparrow$}\\
\cmidrule(lr){2-5}
\cmidrule(lr){6-9}
\cmidrule(lr){10-13}
Model
 & SM & CC & OC & Overall
 & SM & CC & OC & Overall
 & SM & CC & OC & Overall \\
\midrule
\multicolumn{13}{l}{\textit{End-to-end Methods}} \\[2pt]
SAM 3~\cite{carion2025sam}
 & 53.0
 & 61.4
 & -10.5
 & \textbf{38.5}
 & 0.777
 & 0.857
 & -0.241
 & \textbf{0.590}
 & 68.3
 & 71.7
 & 43.8
 & \textbf{65.4}\\
YOLO-World~\cite{cheng2024yolo}
 & 39.4
 & 20.8
 & -5.9
 & 21.1
 & 0.761
 & 0.362
 & -0.296
 & 0.444
 & 51.8
 & 57.6
 & 19.8
 & 47.6 \\
OpenSeeD~\cite{zhang2023simple}
 & 28.9
 & 29.8
 & -4.0
 & 15.1
 & 0.627
 & 0.622
 & -0.613
 & 0.365
 & 46.1
 & 47.9
 & 6.6
 & 41.3 \\
\midrule
\multicolumn{13}{l}{\textit{Multi-model Frameworks}} \\[2pt]
Grounded SAM 2~\cite{ren2024grounded}
 & 13.0
 & 5.9
 & 3.6
 & 9.9
 & 0.217
 & 0.097
 & 0.058
 & 0.165
 & 60.0
 & 60.7
 & 60.8
 & 60.3 \\
OWLv2~\cite{minderer2023scaling} + SAM1~\cite{kirillov2023segment}
 & 43.2
 & 41.0
 & -8.0
 & 27.9
 & 0.845
 & 0.702
 & -0.313
 & 0.564
 & 51.1
 & 58.4
 & 25.6
 & 49.5 \\
\midrule
\multicolumn{13}{l}{\textit{Agentic Methods}} \\[2pt]
CAFE-SAM3 (GPT-5.5)
 & 69.7
 & 66.1
 & 44.7
 & \textbf{63.3 (+24.8)}
 & 0.909
 & 0.877
 & 0.633
 & \textbf{0.843}
 & 76.6
 & 75.3
 & 70.6
 & \textbf{75.1 (+9.7)}\\
\bottomrule
\end{tabular}%
\end{adjustbox}
\label{tab:cafe_concept_results}
\end{table*}

\textbf{Aligned and Unaligned False Positive Rates.} We additionally report the
target-Aligned False Positive Rate (AFPR) and its unaligned counterpart
(UFPR), defined over the full set of negative prompts so they decompose
the standard image-level false positive rate.
Let $N$ denote the total number of paired images, which equals the number
of negative prompts. Following the classification in
Table~\ref{tab:cafe-binclass}, we define
\begin{equation}
\mathrm{AFPR}
=
\frac{\mathrm{TA\text{-}FP}}{N},
\qquad
\mathrm{UFPR}
=
\frac{\mathrm{UA\text{-}FP}}{N},
\end{equation}
where
$N
= \mathrm{TA\text{-}FP}+\mathrm{UA\text{-}FP}+\mathrm{TN}$.
By construction these two rates partition the image-level false positive
rate,
\begin{equation}
\mathrm{IL\text{-}FPR}
=
\frac{\mathrm{TA\text{-}FP}+\mathrm{UA\text{-}FP}}{N}
=
\mathrm{AFPR} + \mathrm{UFPR},
\end{equation}
AFPR isolates the fraction of misleading prompts that produce target-aligned false positives, corresponding to cases where the model assigns high confidence to a semantically invalid query over the edited target region. UFPR captures unaligned false positives, where the misleading prompt elicits a high-confidence response outside the target region.

For each baseline model, we use a calibrated threshold for its presence confidence score. For SAM3, we adopt the default threshold of $0.5$ following its original evaluation protocol. Unless otherwise specified, we report AFPR at an IoU threshold of $\tau=0.3$. A sensitivity analysis with respect to $\tau$ is provided in the Appendix~\ref{appendix:thr_sen}.

\textbf{Concept Swap Rate.} A concept swap occurs when a model loses the original concept on the
target region under $p^{+}$ and simultaneously commits to the
counterfactual concept under $p^{-}$. Following the classification in
Table~\ref{tab:cafe-binclass}, we say $p^{+}$ has \emph{lost} the target
concept whenever it fails to produce a target-aligned true positive,
i.e., $p^{+} \notin \mathrm{TA\text{-}TP}$. We define the
target-Aligned Concept Swap Rate (ACSR) as the joint rate at which
$p^{+}$ loses the concept and $p^{-}$ produces a target-aligned false
positive on the same target region:
\begin{equation}
\mathrm{ACSR}
=
\frac{1}{N}
\sum_{i=1}^{N}
\mathbb{1}
\left[
p_i^{+} \notin \mathrm{TA\text{-}TP}
\;\land\;
p_i^{-} \in \mathrm{TA\text{-}FP}
\right].
\end{equation}
The unaligned counterpart UCSR replaces $\mathrm{TA\text{-}FP}$ with
$\mathrm{UA\text{-}FP}$, capturing concept loss on the target combined
with hallucinated detections elsewhere:
\begin{equation}
\mathrm{UCSR}
=
\frac{1}{N}
\sum_{i=1}^{N}
\mathbb{1}
\left[
y_i^{+} \notin \mathrm{TA\text{-}TP}
\;\land\;
y_i^{-} \in \mathrm{UA\text{-}FP}
\right].
\end{equation}
and the overall Concept Swap Rate decomposes as
\begin{equation}
\mathrm{CSR}
=
\mathrm{ACSR}
+
\mathrm{UCSR}.
\end{equation}
ACSR is the strictest variant, isolating the worst failure mode in
which the counterfactual concept replaces the original on the target
itself; UCSR captures a softer failure where the original concept is
dropped from the target while the counterfactual is hallucinated
elsewhere in the image. 

\section{Experiments}

\begin{table}[!t]
\centering
\caption{
False-positive and concept-swap analysis on CAFE.
FPR measures the proportion of misleading prompts that produce accepted masks, while AFPR reports the target-aligned false-positive rate after excluding unaligned false positives.
ACSR measures the rate at which the positive prompt fails to produce a target-aligned true positive and the misleading negative prompt produces a target-aligned false positive on the same target region.
UFPR and UCSR are the corresponding unaligned counterparts: UFPR counts unaligned false positives under misleading prompts, and UCSR counts cases where the positive prompt loses the target concept while the misleading negative prompt is hallucinated elsewhere in the image.
}
\begin{adjustbox}{width=1\linewidth}
\begin{tabular}{l cccc cccc cccc cc}
\toprule
 & \multicolumn{4}{c}{FPR $\downarrow$}
 & \multicolumn{4}{c}{AFPR $\downarrow$}
 & \multicolumn{4}{c}{ACSR $\downarrow$}
 & \multicolumn{2}{c}{Unconditional $\downarrow$} \\
\cmidrule(lr){2-5}
\cmidrule(lr){6-9}
\cmidrule(lr){10-13}
\cmidrule(lr){14-15}
Model
 & SM & CC & OC & Overall
 & SM & CC & OC & Overall
 & SM & CC & OC & Overall
 & UFPR & UCSR \\
\midrule
\multicolumn{15}{l}{\textit{End-to-end Methods}} \\[2pt]
SAM 3~\cite{carion2025sam}
 & 10.3\%
 & 7.9\%
 & 66.3\%
 & 21.2\%
 & 9.5\%
 & 7.4\%
 & 65.6\%
 & 20.5\%
 & 1.9\%
 & 0.3\%
 & 37.8\%
 & 8.9\%
 & 0.7\%
 & 0.2\%\\
YOLO-World~\cite{cheng2024yolo}
 & 18.7\%
 & 70.5\%
 & 89.6\%
 & 47.6\%
 & 12.5\%
 & 59.2\%
 & 78.1\%
 & 38.9\%
 & 0.9\%
 & 1.5\%
 & 41.6\%
 & 9.5\%
 & 8.7\%
 & 2.6\% \\
OpenSeeD~\cite{zhang2023simple}
 & 1.1\%
 & 4.7\%
 & 63.3\%
 & 14.9\%
 & 0.8\%
 & 3.4\%
 & 62.0\%
 & 14.1\%
 & 0.5\%
 & 0.3\%
 & 58.6\%
 & 12.4\%
 & 0.8\%
 & 0.6\% \\
\midrule
\multicolumn{15}{l}{\textit{Multi-model Frameworks}} \\[2pt]
Grounded SAM 2~\cite{ren2024grounded}
 & 90.5\%
 & 98.1\%
 & 99.3\%
 & 94.5\%
 & 75.7\%
 & 88.0\%
 & 96.4\%
 & 83.4\%
 & 0.8\%
 & 0.5\%
 & 2.5\%
 & 1.1\%
 & 11.1\%
 & 1.8\% \\
OWLv2~\cite{minderer2023scaling} + SAM1~\cite{kirillov2023segment}
 & 7.0\%
 & 25.5\%
 & 62.7\%
 & 23.6\%
 & 4.6\%
 & 19.6\%
 & 60.0\%
 & 20.1\%
 & 1.0\%
 & 1.2\%
 & 48.0\%
 & 10.7\%
 & 3.4\%
 & 0.7\% \\
\midrule
\multicolumn{15}{l}{\textit{Agentic Methods}} \\[2pt]
CAFE-SAM3 (GPT-5.5)
 & 8.1\%
 & 12.0\%
 & 29.2\%
 & \textbf{13.5\%}
 & 7.7\%
 & 9.6\%
 & 25.8\%
 & \textbf{11.9\%}
 & 0.5\%
 & 0.2\%
 & 6.8\%
 & \textbf{1.7\%}
 & 1.6\%
 & 0.1\% \\
\bottomrule
\end{tabular}%
\end{adjustbox}
\label{tab:cafe_fp_swap_results}
\end{table}

\subsection{Results on Segmentation Models and Modular Frameworks.} We evaluate end-to-end open-vocabulary segmentation models, modular frameworks combining open-vocabulary detectors with SAM~\cite{kirillov2023segment}/SAM2~\cite{ravi2024sam}, and agentic methods that perform explicit verification using SAM3. Baseline details are in Appendix~\ref{app:model_versions}. Table~\ref{tab:cafe_concept_results} reports cgF$_1$, IL\_MCC, and pmF$_1$ on CAFE. Non-agentic models maintain relatively high pmF$_1$, indicating positive prompts can still be localized. However, low IL\_MCC and cgF$_1$ highlight that the core challenge is rejecting semantically invalid concepts, not positive-case segmentation. Grounded SAM2 illustrates this: stable pmF$_1$ across SM, CC, and OC coexists with consistently low IL\_MCC, showing that accurate masks do not guarantee semantic-validity judgments. OC is the most difficult category. Most non-agentic models achieve negative IL\_MCC on OC, revealing inverse correlation with semantic labels. Even SAM3, despite an image-level presence head and strong overall performance, drops from 0.857 IL\_MCC on CC to -0.241 on OC, suggesting that presence prediction alone cannot resolve ontological counterfactuals. Table~\ref{tab:cafe_fp_swap_results} analyzes false positives (FPR) and concept swaps. First, non-agentic models exhibit high FPR, frequently accepting misleading prompts. Second, most false positives are target-aligned (IoU $>0.3$), except for YOLO-World and Grounded SAM2, indicating that counterfactually edited regions drive errors. Third, OC shows the highest FPR and AFPR across models. In extreme cases, a model may accept a misleading prompt while rejecting the positive one, reflected in ACSR. Grounded SAM2's low ACSR results from accepting both positive and misleading prompts rather than robust rejection, and must be interpreted alongside its high FPR and AFPR.

Overall, these results demonstrate that current open-vocabulary segmentation models struggle to distinguish sculptural or artificial depictions from living entities under ontological conflict, and that positive mask quality does not imply reliable semantic grounding.

\subsection{Does Explicit Reasoning Help Counterfactual Segmentation?} Leveraging the strong understanding ability of current VLMs for various type of visual content~\cite{wang2025videoverse, wang2026towards, wei2025tiif}, CAFE-SAM3 agent (GPT-5.5) demonstrates the benefit of agentic verification, with details in Appendix~\ref{app:SAM3-CAFE Agent}. Compared with direct SAM3, overall cgF$_1$ rises from 38.5 to 63.3, IL\_MCC from 0.590 to 0.843, and pmF$_1$ from 65.4 to 75.1. The largest gains occur on OC, with cgF$_1$ increasing from -10.5 to 44.7 and IL\_MCC from -0.241 to 0.633, highlighting the utility of explicit reasoning when distinguishing semantically valid concepts from visually plausible but ontologically invalid cues. False-positive also shows that, CAFE-SAM3 reduces overall FPR from 21.2\% to 13.5\%, AFPR from 20.5\% to 11.9\%, and ACSR from 8.9\% to 1.7\% compared with SAM3. Gains are especially pronounced on OC (FPR 66.3\% $\to$ 29.2\%, AFPR 65.6\%$\to$25.8\%, ACSR 37.8\%$\to$6.8\%), indicating that agentic verification primarily improves rejection of semantically invalid target-aligned masks rather than positive-case segmentation. OC false-positive rates remain higher than those of SM and CC, suggesting room for improvement in handling ontological counterfactuals. Two additional insights emerge. First, SAM3's image-level presence head enhances robustness on SM and CC, where misleading cues stem from surface appearance or context, but is insufficient for OC. Second, the strong improvement of CAFE-SAM3 agent shows that explicit verification enables segmentation systems to better distinguish semantically invalid concepts from visually plausible counterfactual cues.

\section{Conclusion}

We introduced CAFE, a counterfactual attribute factuality evaluation framework for promptable concept segmentation, comprising 2,146 paired samples with positive and misleading prompts across Superficial Mimicry, Context Conflict, and Ontological Conflict. Our results reveal that current open-vocabulary segmentation models often fail to reject semantically invalid concepts under counterfactual cues. SAM3's image-level presence head improves robustness in some cases but remains insufficient for ontological conflicts. CAFE-SAM3 agent demonstrates that MLLM-based reasoning can reduce false positives and concept swaps, suggesting a path toward more reliable promptable segmentation.

\section{Limitations} CAFE currently evaluates a single counterfactually edited target per image, allowing controlled assessment of misleading prompts. It does not cover more complex scenes with multiple counterfactual instances or co-occurrence with unedited instances of the same or related concepts. Consequently, counterfactual robustness in crowded or mixed-instance scenarios remains untested.

\newpage
\clearpage
\appendix
\appendix

\section{Dataset Preparation}
\label{appendix:dataset_gen}
\subsection{CAFE Annotation Pipeline}
\begin{figure}[!b]
  \centering
  
  \includegraphics[width=1.0\textwidth]{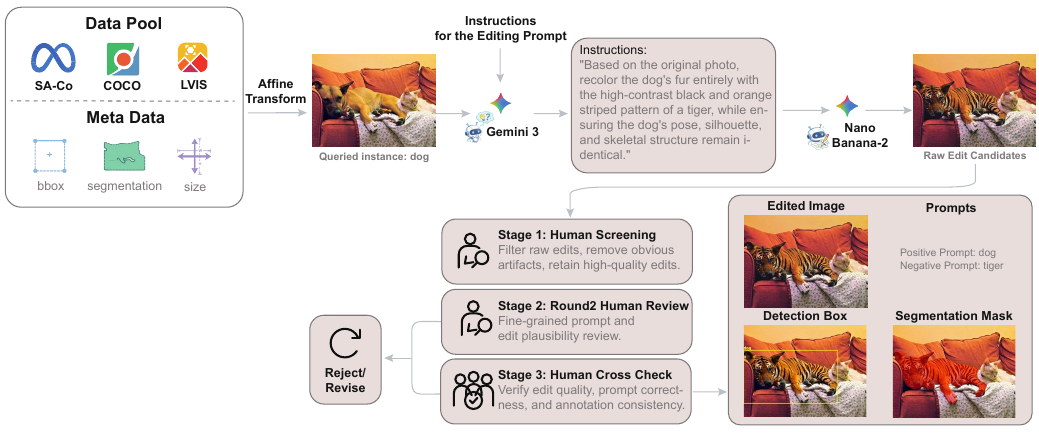}
  \caption{Overview of the CAFE dataset annotation pipeline. We draw image-annotation pairs from COCO, SA-Co, and LVIS. The images and annotations are first processed with affine transformations to fit the input size required by Gemini 3, and are then fed into Gemini 3 to generate corresponding editing instructions. Based on the generated instructions, we use nano-banana to perform image editing for all three counterfactual categories. The raw edits then undergo a three-stage filtering and cross-checking process. In the first stage, human annotators filter the raw edits and remove images with obvious artifacts. In the second stage, human annotators perform a fine-grained review of edit quality and prompt plausibility for both positive and negative prompts. In the third stage, three human editors cross-check all remaining pairs and produce the final high-quality dataset.}
  \label{fig:cafe_pipeline}
\end{figure}

\begin{figure*}[!t]
    \centering
    \includegraphics[width=\textwidth]{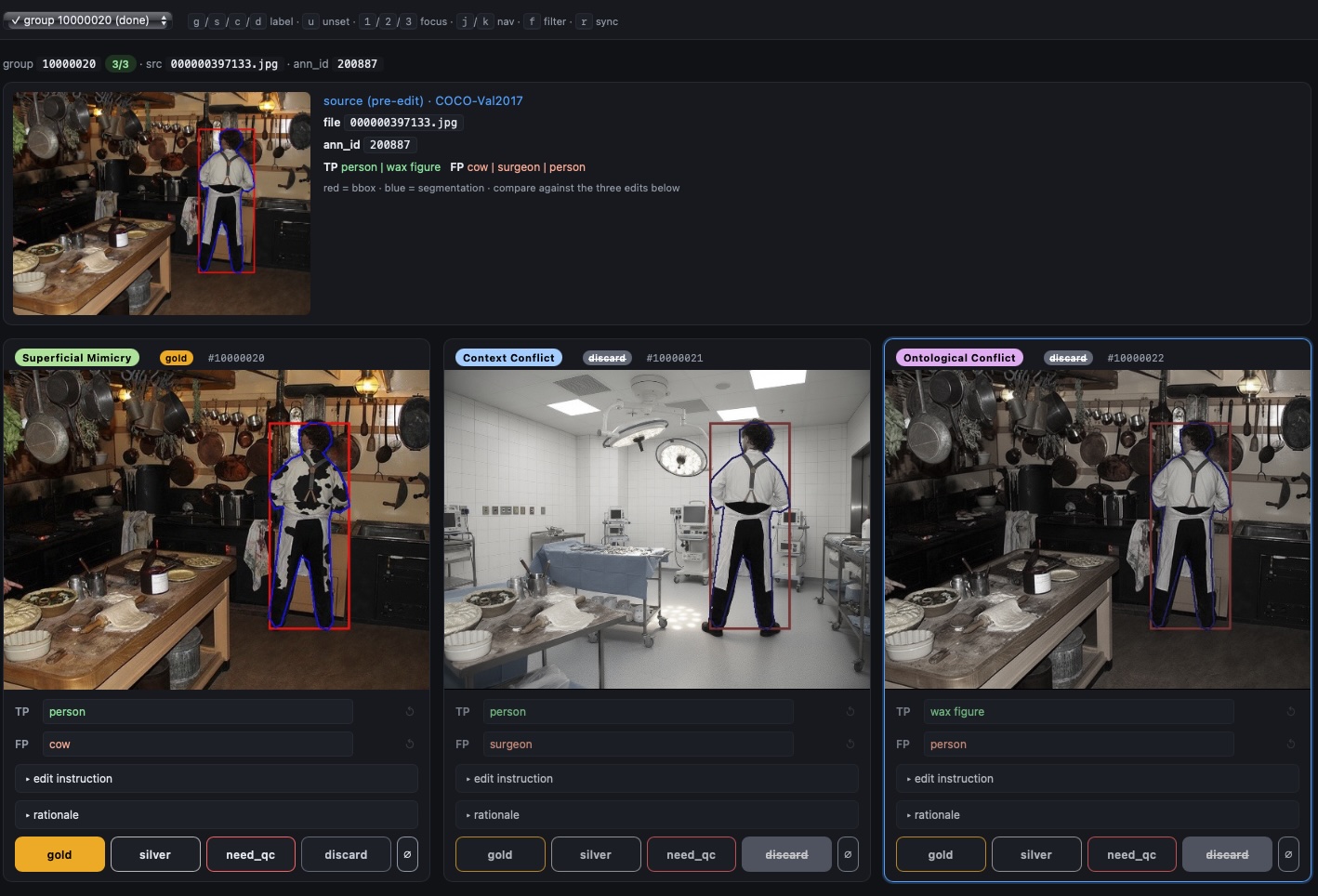}
    \caption{
    Data annotation engine used for human quality inspection. Human annotators use the interface to check edit plausibility, mask alignment, and prompt validity during the multi-round filtering process.
    }
    \label{fig:frontend}
\end{figure*}

The CAFE annotation pipeline is shown in Fig.~\ref{fig:cafe_pipeline}. To fit the input resolution of Gemini, we apply affine transformations to the original images and annotations from the validation sets of COCO~\cite{lin2014microsoft}, LVIS~\cite{gupta2019lvis}, and SA-Co/Gold~\cite{carion2025sam}. The transformed annotations are inherited from the source image-annotation pairs, while Gemini-3 is used to generate editing instructions with prompt-engineered inputs containing multiple in-context cases based on the queried instance and the input image. Details of the prompt-engineering cases are provided in Appendix~\ref{p:editing_prompts}.
Based on the generated editing instructions, Nano-banana-2 performs image editing. The annotation format includes the positive prompt, the negative prompt, the corresponding editing type, the editing instruction, and the rationale. The 48,423 raw generated samples are then filtered by human annotators to remove low-quality cases, including poor mask alignment and implausible editing instructions. The filtered images are further reviewed by three human experts. An image is included in the final dataset only when at least two reviewers agree that the edit is reliable and semantically valid, thereby reducing the effect of individual annotator bias. The interface of the annotation frontend is shown in Fig.~\ref{fig:frontend}. After this high-selectivity filtering process, 2,146 samples are retained for the final dataset, corresponding to a retention rate of 4.4\%.

\subsection{Prompts and Models for Dataset Generation}
\label{p:editing_prompts}

\subsubsection{Shared Task Head and Output Schema}

\label{p:task_head}

\lstdefinelanguage{XMLLite}{
  morestring=[b]",
  morecomment=[s]{<!--}{-->},
  morekeywords={<type>,</type>,<edit_instruction>,</edit_instruction>,<rationales>,</rationales>,<positive_prompt>,</positive_prompt>,<negative_prompt>,</negative_prompt>}
}
\lstset{
  basicstyle=\ttfamily\small,
  columns=fullflexible,
  breaklines=true,
  showstringspaces=false,
  frame=none,
  xleftmargin=0pt,
  tabsize=2,
  keywordstyle=\color{black!75}, 
  commentstyle=\color{black!55},
}

\begin{tcolorbox}[
  breakable,
  colback=gray!8,
  colframe=gray!70!black,
  coltitle=white,
  colbacktitle=gray!70!black,
  title=\textbf{CAFE Auto-Prompt Pipeline -- Shared Header \& Output Schema},
  fonttitle=\bfseries,
  boxrule=0.8pt, arc=3pt,
  left=8pt, right=8pt, top=8pt, bottom=8pt,
  titlerule=0pt,
  separator sign=\textcolor{gray!70!black}{\rule{\linewidth}{0.6pt}},
  before upper={\setlength{\parskip}{4pt}},
]
\textbf{I.~Task Overview.}\;
You MUST generate a \texttt{<TARGET\_TYPE>} counterfactual edit for the
\texttt{<INSTANCE\_CATEGORY>} instance in the given image. Do \emph{not}
produce any other edit type. If no valid edit of this type exists for this
instance, return exactly the literal string \texttt{None} and terminate.

\tcblower

\textbf{II.~Output Format.}\;
The model must return \textbf{exactly} the following XML-style fields.

\begin{tcolorbox}[
  enhanced,
  colback=gray!3,                
  colframe=gray!55!black,      
  boxrule=0.5pt,
  arc=2pt,
  left=6pt,right=6pt,top=6pt,bottom=6pt
]
\begin{lstlisting}[language=XMLLite]
<type>
[The edit type. Must be exactly the target type specified above.]
</type>

<edit_instruction>
[One-sentence, imperative editing directive. Specify exactly what changes and explicitly state what must remain unchanged.]
</edit_instruction>

<rationales>
[One or two sentences explaining the intended conflict. No repetition of the instruction.]
</rationales>

<positive_prompt>
[For Superficial Mimicry and Context Conflict, the positive prompt is the original category of the input instance. For Ontological Conflict, it is the category of the newly rendered instance.]
</positive_prompt>

<negative_prompt>
[For Superficial Mimicry, the negative prompt is the category being mimicked. For Context Conflict, it must be a concrete object category plausible in the new context but distinct from the original instance. For Ontological Conflict, it is the category of the original instance. In all cases, the negative prompt must be a standard, atomic object category. Context-dependent phrases like "street animal", "snow animal", "beach dog" are strictly prohibited.]
</negative_prompt>
\end{lstlisting}
\end{tcolorbox}
\end{tcolorbox}

\subsubsection{Prompt for Superficial Mimicry Editing Instruction Generation}
\label{p:SM}
\begin{tcolorbox}[
  breakable,
  colback=cafemim!6,
  colframe=cafemim!75!black,
  coltitle=white,
  colbacktitle=cafemim!75!black,
  title=\textbf{I.\;Rules for \emph{Superficial Mimicry}},
  fonttitle=\bfseries,
  boxrule=0.8pt, arc=3pt,
  left=8pt, right=8pt, top=8pt, bottom=8pt, titlerule=0pt
]
\textbf{Surface-Only Change.}
\begin{itemize}\setlength\itemsep{1pt}
  \item Modification must affect appearance only (color, pattern, or printed texture).
  \item The underlying material, volume, thickness, and physical structure must remain unchanged.
  \item The edit must behave like a thin visual overlay (paint, print, dye, wrapping), \emph{not} a material transformation.
  \item The applied appearance must preserve overall identity recognizability of the object category.
  \item The conflicting concept must exist as a physical surface appearance, not as a depicted image (screen content, photo, poster, logo).
  \item Appearances implying volume-level material transformation (solid metal body, glass body, stone sculpture) are not allowed.
  \item Textures or patterns without a distinctive, widely recognizable category-level signature are not allowed.
  \item The applied appearance must introduce visual cues strongly associated with a different object category.
\end{itemize}

\textbf{Global Coverage.}
\begin{itemize}\setlength\itemsep{1pt}
  \item The new appearance must uniformly cover the entire visible surface of the instance.
  \item All visible parts and components of the object must be consistently modified.
  \item Partial edits, localized patches, or unchanged regions are not allowed.
\end{itemize}

\textbf{Label Exclusivity.}
\begin{itemize}\setlength\itemsep{1pt}
  \item The \texttt{negative prompt} must represent a different object category, not a material variant, subtype, or attribute description of the same object.
\end{itemize}
\end{tcolorbox}

\subsubsection{Prompt for Ontological Conflict Editing Instruction Generation}
\label{p:OC}
\begin{tcolorbox}[
  breakable,
  colback=cafeont!6,
  colframe=cafeont!75!black,
  coltitle=white,
  colbacktitle=cafeont!75!black,
  title=\textbf{II.\;Rules for \emph{Ontological Conflict}},
  fonttitle=\bfseries,
  boxrule=0.8pt, arc=3pt,
  left=8pt, right=8pt, top=8pt, bottom=8pt, titlerule=0pt
]
\textbf{Substance Replacement.}
\begin{itemize}\setlength\itemsep{1pt}
  \item The object must appear to be entirely made of a different substance while its geometry is unchanged.
  \item The transformation must change the object's physical nature (living $\to$ inanimate, organic $\to$ mineral, functional $\to$ sculpture).
  \item The transformation must imply a new object identity (sculpture, statue, mineral object, fluid mass), not a material variant of the original object.
\end{itemize}

\textbf{Complete Transformation.}
\begin{itemize}\setlength\itemsep{1pt}
  \item All visible parts must share the new substance.
  \item Partial material changes (only surface, only fur, or selected components) are not allowed.
\end{itemize}

\textbf{Volume-Level Change.}
\begin{itemize}\setlength\itemsep{1pt}
  \item The transformation must imply a full volumetric material change, not a surface coating, paint, or texture overlay.
\end{itemize}

\textbf{Category Displacement.}
\begin{itemize}\setlength\itemsep{1pt}
  \item If the original category is defined by its function (fork, bicycle), the transformation must render the object non-functional (a transparent soap-film fork, a stone statue of a bicycle).
  \item After the transformation the object must be more naturally interpreted as the new category than the original. If global geometry still strongly determines the original category (ice table, wooden bicycle, marble toilet), the edit is invalid.
\end{itemize}

\textbf{Semantic Exclusivity.}
\begin{itemize}\setlength\itemsep{1pt}
  \item If the edited object would still be naturally described as ``a [material] + original category'', the edit is invalid. The original category must no longer serve as the primary identity.
\end{itemize}
\end{tcolorbox}

\subsubsection{Prompt for Context Conflict Editing Instruction Generation}
\label{p:CC}
\begin{tcolorbox}[
  breakable,
  colback=cafectx!6,
  colframe=cafectx!75!black,
  coltitle=white,
  colbacktitle=cafectx!75!black,
  title=\textbf{III.\;Rules for \emph{Context Conflict}},
  fonttitle=\bfseries,
  boxrule=0.8pt, arc=3pt,
  left=8pt, right=8pt, top=8pt, bottom=8pt, titlerule=0pt
]
\textbf{Foreground Invariance.}
\begin{itemize}\setlength\itemsep{1pt}
  \item Target instance pixels must remain \emph{identical} to the original image.
  \item No recoloring, relighting, shadow change, specular change, reflection change, refraction change, blur change, edge halo, or color spill on the instance.
  \item No new occluders may overlap the instance. No transparency changes, no ``see-through'' effects.
  \item Any change inside the instance mask invalidates the sample.
\end{itemize}

\textbf{Background-Only Edit.}
\begin{itemize}\setlength\itemsep{1pt}
  \item Only pixels outside the instance mask may change.
  \item Replace the environment with a photo-realistic background that matches the original camera viewpoint, perspective, and depth-of-field, but must not alter the instance appearance.
  \item Do not add, remove, or modify any foreground objects in a way that could plausibly create competing instances of the negative-prompt category.
\end{itemize}

\textbf{Counterfactual Strength.}
\begin{itemize}\setlength\itemsep{1pt}
  \item The new environment must be strongly incompatible with the instance's typical context (an ``implausible placement''), not a common co-occurrence.
  \item If the new environment is a natural or frequent context for the object, the sample is invalid.
\end{itemize}
\end{tcolorbox}


\subsubsection{In-context cases for Superficial Mimicry}
\label{ctx:SM}
\begin{tcolorbox}[
  breakable,
  colback=cafemim!4,
  colframe=cafemim!70!black,
  coltitle=white,
  colbacktitle=cafemim!70!black,
  title=\textbf{In-Context Cases -- \emph{Superficial Mimicry}},
  fonttitle=\bfseries,
  boxrule=0.6pt, arc=3pt,
  left=8pt, right=8pt, top=6pt, bottom=6pt, titlerule=0pt
]
\textbf{(A) Positive Cases}

\vspace{0.3em}
\textit{Case A1.}\;\emph{Input instance: elephant.}
\begin{itemize}\setlength\itemsep{1pt}
  \item \texttt{<edit\_instruction>} Recolor the elephant's skin with the pattern of a zebra. Keep the elephant's pose, shape, and lighting identical; the entire background must be unchanged.
  \item \texttt{<rationales>} The elephant is body-painted with zebra-like stripes; large animals can be temporarily painted for exhibition or filming, so the photo is physically plausible while creating a surface-only conflict between the zebra's iconic striping and the elephant's intact anatomy.
  \item \texttt{<Positive Prompt>} elephant \quad \texttt{<Negative Prompt>} zebra
\end{itemize}

\vspace{0.3em}
\textit{Case A2.}\;\emph{Input instance: chair.}
\begin{itemize}\setlength\itemsep{1pt}
  \item \texttt{<edit\_instruction>} Overlay the entire chair with a continuous coral reef texture of branching pink and white corals, preserving every contour, leg, and seat surface of the chair unchanged.
  \item \texttt{<rationales>} Surface decals or spray-painted murals can coat furniture with biological motifs, so a coral-patterned chair is a realistic installation-art prop while keeping its ontology as a chair intact.
  \item \texttt{<Positive Prompt>} chair \quad \texttt{<Negative Prompt>} coral
\end{itemize}

\vspace{0.3em}
\textit{Case A3.}\;\emph{Input instance: laptop.}
\begin{itemize}\setlength\itemsep{1pt}
  \item \texttt{<edit\_instruction>} Apply a photorealistic ladybug shell texture (red with black polka dots and a glossy chitinous finish) uniformly across the laptop's case, preserving ports, keyboard, hinge, and all geometric details.
  \item \texttt{<rationales>} Laptop skins in printed decal form are commercially available, so a ladybug-shell laptop is a realistic product staging while producing a pure textural conflict.
  \item \texttt{<Positive Prompt>} laptop \quad \texttt{<Negative Prompt>} ladybug
\end{itemize}

\vspace{0.6em}
\textbf{(B) Negative Cases}

\vspace{0.3em}
\textit{Case B1.}\;\emph{Input instance: potted plant.}
\begin{itemize}\setlength\itemsep{1pt}
  \item \texttt{<edit\_instruction>} Change the plant's leaves and stem texture to resemble metallic wires and circuits, while keeping its silhouette and spatial extent unchanged.
  \item \emph{Why invalid:} replacing leaves and stems with metal circuits implies a material-level transformation rather than a surface overlay, which crosses into Ontological-Conflict territory (\textbf{Surface-Only Change violated}).
  \item \texttt{<Positive Prompt>} potted plant \quad \texttt{<Negative Prompt>} circuits
\end{itemize}

\vspace{0.3em}
\textit{Case B2.}\;\emph{Input instance: TV.}
\begin{itemize}\setlength\itemsep{1pt}
  \item \texttt{<edit\_instruction>} Change the TV screen display to show an aquarium scene with vivid corals and colorful fish, keeping the TV's shape, frame, and positioning identical and the background unchanged.
  \item \emph{Why invalid:} the aquarium exists only as depicted imagery on the screen, not as a distinct physical surface, so it is not a valid Superficial-Mimicry case (\textbf{Conflicting concept must be a physical surface, not depicted content}).
  \item \texttt{<Positive Prompt>} TV \quad \texttt{<Negative Prompt>} aquarium
\end{itemize}

\vspace{0.3em}
\textit{Case B3.}\;\emph{Input instance: bench.}
\begin{itemize}\setlength\itemsep{1pt}
  \item \texttt{<edit\_instruction>} Apply a realistic wooden texture to the bench as a surface coating, while keeping its silhouette, volume, and pose identical.
  \item \emph{Why invalid:} the negative prompt ``wooden bench'' is a true-positive refinement of the original object, not a distinct confounding class (\textbf{Label Exclusivity violated}).
  \item \texttt{<Positive Prompt>} bench \quad \texttt{<Negative Prompt>} wooden bench
\end{itemize}
\end{tcolorbox}

\subsubsection{In-context cases for Ontological Conflict}
\label{ctx:OC}
\begin{tcolorbox}[
  breakable,
  colback=cafeont!4,
  colframe=cafeont!70!black,
  coltitle=white,
  colbacktitle=cafeont!70!black,
  title=\textbf{In-Context Cases -- \emph{Ontological Conflict}},
  fonttitle=\bfseries,
  boxrule=0.6pt, arc=3pt,
  left=8pt, right=8pt, top=6pt, bottom=6pt, titlerule=0pt
]
\textbf{(A) Positive Cases}

\vspace{0.3em}
\textit{Case A1.}\;\emph{Input instance: dog.}
\begin{itemize}\setlength\itemsep{1pt}
  \item \texttt{<edit\_instruction>} Transform the dog's entire body into a translucent crystal sculpture; preserve the skeletal pose, muscular silhouette, and fur-like surface micro-geometry through refractive and reflective glass-like properties; keep the background unchanged.
  \item \texttt{<rationales>} Life-sized crystal animal sculptures are a recognized art form, so converting a dog into a glass-cast form is physically realizable and establishes a new primary identity as sculpture.
  \item \texttt{<Positive Prompt>} crystal sculpture \quad \texttt{<Negative Prompt>} living dog
\end{itemize}

\vspace{0.3em}
\textit{Case A2.}\;\emph{Input instance: bicycle.}
\begin{itemize}\setlength\itemsep{1pt}
  \item \texttt{<edit\_instruction>} Transform the bicycle into a life-size bronze monument-style statue, preserving every tube, wheel, and handlebar shape, with a realistic patinated bronze surface; keep the background identical.
  \item \texttt{<rationales>} Bronze monuments of everyday vehicles exist in sculpture parks, so casting a bicycle in bronze is realistic; the monument identity supplants the functional bicycle identity.
  \item \texttt{<Positive Prompt>} bronze sculpture \quad \texttt{<Negative Prompt>} functional bicycle
\end{itemize}

\vspace{0.3em}
\textit{Case A3.}\;\emph{Input instance: cake.}
\begin{itemize}\setlength\itemsep{1pt}
  \item \texttt{<edit\_instruction>} Transform the cake into a fully polished granite monolith, preserving the tiered geometry, edge details, and spatial orientation; keep the background unchanged.
  \item \texttt{<rationales>} Granite sculptures of everyday objects are a known sculptural trope; the substance change from sponge/cream to granite supplants the cake identity with a stone-monument identity.
  \item \texttt{<Positive Prompt>} granite sculpture \quad \texttt{<Negative Prompt>} edible cake
\end{itemize}

\vspace{0.6em}
\textbf{(B) Negative Cases}

\vspace{0.3em}
\textit{Case B1.}\;\emph{Input instance: cat.}
\begin{itemize}\setlength\itemsep{1pt}
  \item \texttt{<edit\_instruction>} Transform the cat's fur into a solid, glossy marble texture while keeping its pose, shape, and spatial extent unchanged.
  \item \emph{Why invalid:} the instruction limits the transformation to fur, leaving other body parts biological -- a partial substance change does not constitute a full ontological conflict (\textbf{Complete Transformation violated}).
  \item \texttt{<Positive Prompt>} marble sculpture \quad \texttt{<Negative Prompt>} living cat
\end{itemize}

\vspace{0.3em}
\textit{Case B2.}\;\emph{Input instance: dining table.}
\begin{itemize}\setlength\itemsep{1pt}
  \item \texttt{<edit\_instruction>} Transform the dining table's substance into transparent ice while preserving its original shape, dimensions, and spatial orientation.
  \item \emph{Why invalid:} an ice dining table is still functionally a dining table, so ``dining table'' remains a correct label and the transformation fails the label-exclusivity requirement (\textbf{Category Displacement / Semantic Exclusivity violated}).
  \item \texttt{<Positive Prompt>} ice sculpture \quad \texttt{<Negative Prompt>} dining table
\end{itemize}

\vspace{0.3em}
\textit{Case B3.}\;\emph{Input instance: toilet.}
\begin{itemize}\setlength\itemsep{1pt}
  \item \texttt{<edit\_instruction>} Transform the toilet into solid marble while preserving its original shape, size, and positioning; keep the background completely unchanged.
  \item \emph{Why invalid:} marble toilets are a real luxury product and are still semantically toilets, so the test fails the requirement that the original category be supplanted (\textbf{Semantic Exclusivity violated}).
  \item \texttt{<Positive Prompt>} marble sculpture \quad \texttt{<Negative Prompt>} toilet
\end{itemize}
\end{tcolorbox}

\subsubsection{In-context cases for Context Conflict}
\label{ctx:CC}

\begin{tcolorbox}[
  breakable,
  colback=cafectx!4,
  colframe=cafectx!70!black,
  coltitle=white,
  colbacktitle=cafectx!70!black,
  title=\textbf{In-Context Cases -- \emph{Context Conflict}},
  fonttitle=\bfseries,
  boxrule=0.6pt, arc=3pt,
  left=8pt, right=8pt, top=6pt, bottom=6pt, titlerule=0pt
]
\textbf{(A) Positive Cases}

\vspace{0.3em}
\textit{Case A1.}\;\emph{Input instance: polar bear.}
\begin{itemize}\setlength\itemsep{1pt}
  \item \texttt{<edit\_instruction>} Replace the arctic background with a dense, misty bamboo forest. The polar bear instance pixels must remain strictly identical; do not introduce shadows, reflections, occlusions, edge halos, or any pandas/animals.
  \item \texttt{<rationales>} The polar bear is placed in a panda's iconic bamboo habitat, forcing the model to rely on the bear's actual coloration rather than the habitat prior.
  \item \texttt{<Positive Prompt>} polar bear \quad \texttt{<Negative Prompt>} panda
\end{itemize}

\vspace{0.3em}
\textit{Case A2.}\;\emph{Input instance: dog.}
\begin{itemize}\setlength\itemsep{1pt}
  \item \texttt{<edit\_instruction>} Replace the background with an underwater reef scene while preserving the dog's pixels, pose, lighting, and body contours exactly. Do not introduce fish or sharks; the underwater lighting context must not spill onto the dog's body.
  \item \texttt{<rationales>} A land mammal placed underwater tests robustness against marine-mammal priors; without geometric overlap, the model should still recognize the dog from its anatomy.
  \item \texttt{<Positive Prompt>} dog \quad \texttt{<Negative Prompt>} seal
\end{itemize}

\vspace{0.3em}
\textit{Case A3 (vehicle alternative).}\;\emph{Input instance: boat.}
\begin{itemize}\setlength\itemsep{1pt}
  \item \texttt{<edit\_instruction>} Replace the water around the boat with a dry desert landscape of rolling dunes, keeping the boat's hull, sails, and rigging pixels identical to the original image. Do not add any sand vehicles.
  \item \texttt{<rationales>} A seaworthy boat stranded on dunes tests habitat-driven confusion with sand sleds while the boat's geometry remains intact.
  \item \texttt{<Positive Prompt>} boat \quad \texttt{<Negative Prompt>} sand sled
\end{itemize}

\vspace{0.6em}
\textbf{(B) Negative Cases}

\vspace{0.3em}
\textit{Case B1.}\;\emph{Input instance: cat.}
\begin{itemize}\setlength\itemsep{1pt}
  \item \texttt{<edit\_instruction>} Replace the bathroom environment with a bustling city street, while keeping the cat's pixels, pose, and lighting exactly identical.
  \item \emph{Why invalid:} a cat on a city street is a common, natural occurrence, so the context is not counterfactual enough; the negative prompt ``street animal'' is also a hypernym rather than an atomic category (\textbf{Counterfactual Strength violated}).
  \item \texttt{<Positive Prompt>} cat \quad \texttt{<Negative Prompt>} street animal
\end{itemize}

\vspace{0.3em}
\textit{Case B2.}\;\emph{Input instance: bed.}
\begin{itemize}\setlength\itemsep{1pt}
  \item \texttt{<edit\_instruction>} Replace the background with a stadium full of cheering fans while keeping the bed and all objects on it unchanged.
  \item \emph{Why invalid:} ``stadium'' describes a scene/place, not a competing foreground object, so it cannot be a valid negative prompt for a bed (\textbf{Counterfactual Strength violated}).
  \item \texttt{<Positive Prompt>} bed \quad \texttt{<Negative Prompt>} stadium
\end{itemize}

\vspace{0.3em}
\textit{Case B3.}\;\emph{Input instance: sink.}
\begin{itemize}\setlength\itemsep{1pt}
  \item \texttt{<edit\_instruction>} Replace the bathroom background with a realistic outdoor beach setting, while keeping the sink pixels unchanged.
  \item \emph{Why invalid:} ``beach equipment'' is a functional collection noun without a definable geometry, so it cannot be a valid segmentation-level negative prompt (\textbf{Counterfactual Strength violated}).
  \item \texttt{<Positive Prompt>} sink \quad \texttt{<Negative Prompt>} beach equipment
\end{itemize}
\end{tcolorbox}

\subsubsection{More Discussions on the Prompts for Ontological Conflict}

In this section, we discuss how we design the positive and negative prompts for ontological conflicts.

We are aware that in earlier benchmarks such as LVIS~\cite{gupta2019lvis} and SA-Co~\cite{carion2025sam}, an instance may correspond to multiple positive categories. For example, LVIS~\cite{gupta2019lvis} emphasizes the annotation of overlapping categories: a toy deer can be annotated as a toy, a deer, and a toy deer. Such ambiguity is acceptable in earlier datasets, since language ambiguity naturally exists in category annotation. However, for counterfactual reasoning and scenarios that require concept-faithful grounding, more exclusive and precise referring expressions are needed.

\begin{figure}[t]
    \centering
    \includegraphics[width=\linewidth]{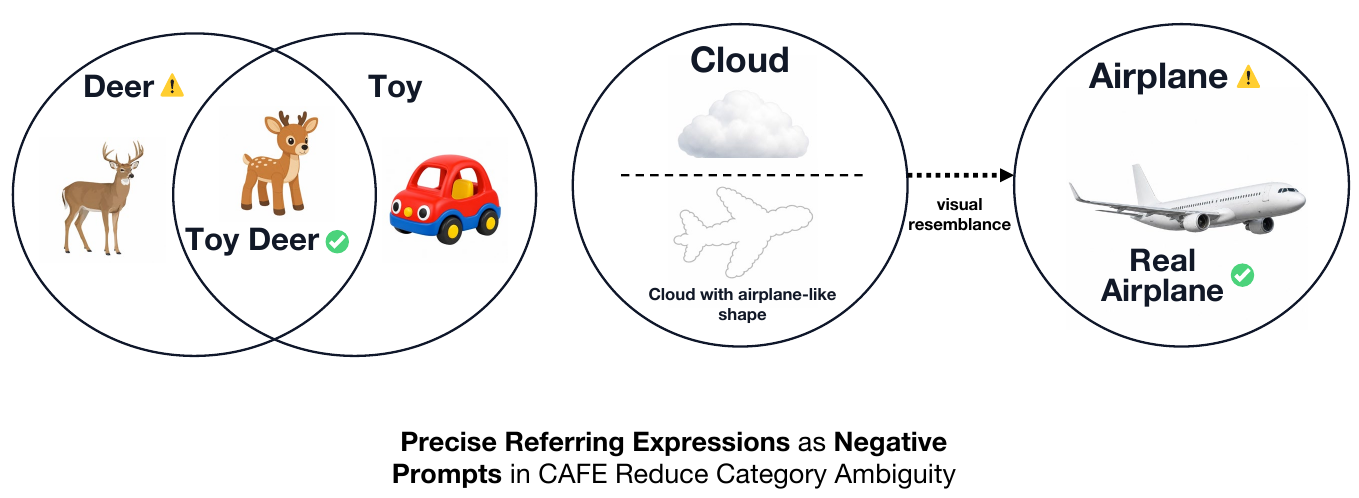}
    \caption{
    Illustration of category ambiguity and prompt disambiguation in ontological conflicts.
    Left: an object may validly belong to multiple categories, such as a toy deer belonging to both the toy and deer categories.
    Right: a cloud with an airplane-like shape visually resembles an airplane, but it remains a cloud rather than a real airplane.
    Therefore, CAFE uses precise negative prompts such as ``real airplane'' instead of ambiguous prompts such as ``airplane'' to reduce semantic ambiguity.
    }
    \label{fig:ontological-prompt-discussion}
\end{figure}

To this end, and to avoid controversial cases, all negative prompts in the ontological conflict category are strictly verified by human expert annotators and cross-checked based on consensus. These negative prompts are constructed with restrictive modifiers to reduce semantic ambiguity, as illustrated in Fig.~\ref{fig:ontological-prompt-discussion}. For example, if the original object is an airplane but the edited target region is re-rendered as a cloud with an airplane-like shape, we use ``real airplane'' as the negative prompt instead of the standalone prompt ``airplane''. This avoids the ambiguity caused by visual resemblance between the edited cloud region and the original object category. Similarly, when a person is re-rendered as a sculpture, we use prompts such as ``living human'' or ``real person'' as the negative prompts to avoid overlap with ambiguous categories. The same principle applies to other objects: if a blender is re-rendered as a wax sculpture and human consensus determines that the edited target is a sculpture, we use ``functional blender'' rather than the standalone prompt ``blender'' as the negative prompt.

This design ensures that the ontological conflict cases can test model hallucination with minimal semantic controversy. All ontological conflict edits are strictly reviewed, and their accepted proportion is therefore lower than that of the other two categories, since the acceptance criteria are intentionally stringent.

\clearpage
\section{More Examples from CAFE}
\label{app:more_examples}

\begin{figure*}[h]
\centering
\includegraphics[width=0.95\textwidth]{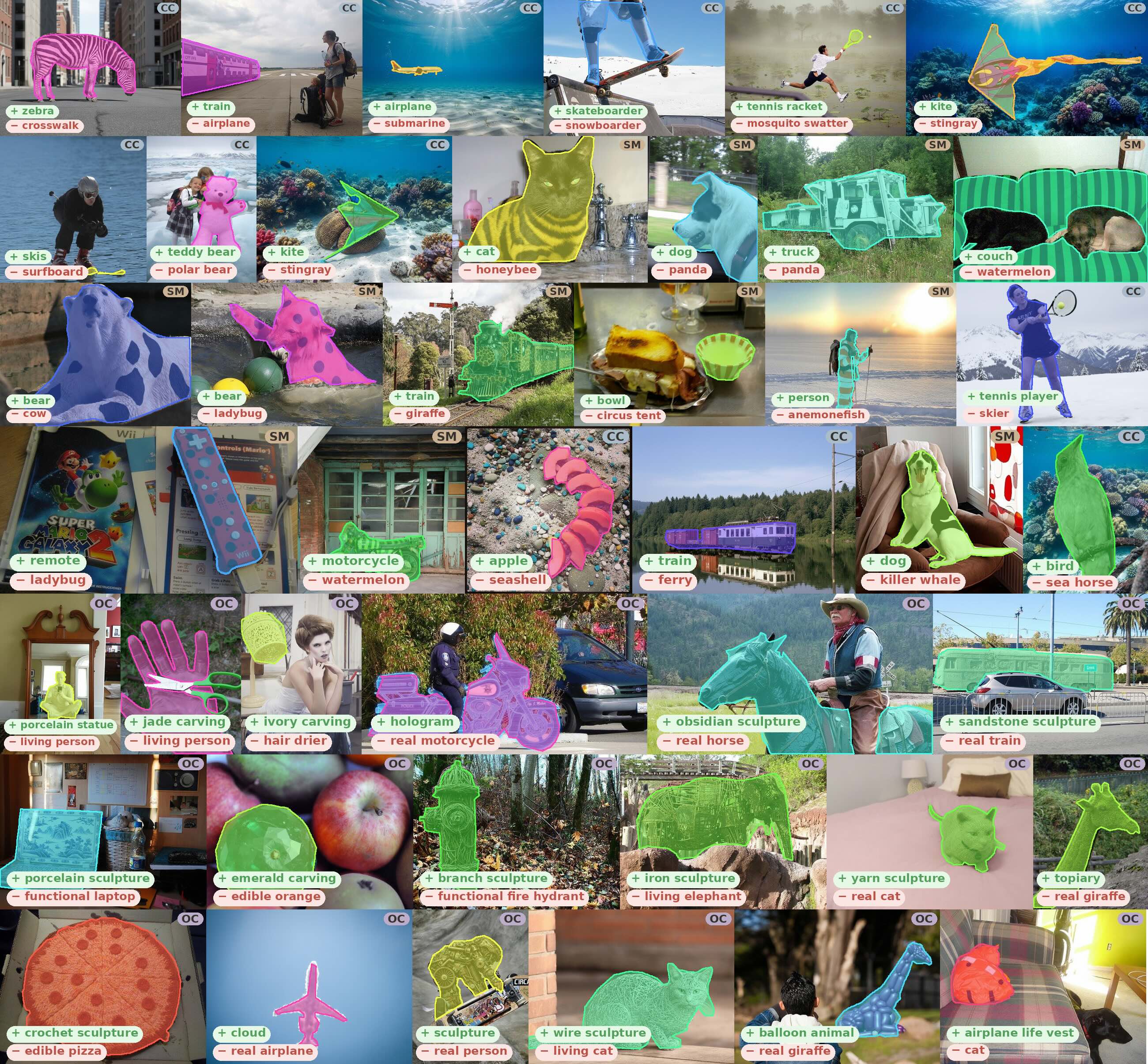}
\caption{
Additional examples from CAFE. 
Each sample consists of a counterfactually edited image, an inherited target mask, a semantically valid positive prompt, and a visually plausible but semantically invalid misleading prompt. 
The examples cover Superficial Mimicry (SM), Context Conflict (CC), and Ontological Conflict (OC), demonstrating the diversity of object categories, prompt pairs, and attribute-level counterfactual conflicts in CAFE.
}
\label{fig:more_cafe_examples}
\end{figure*}

\clearpage

\section{Implementation Details for Baseline Model Evaluation}

\subsection{Compute Resources.}
All experiments are evaluation-only inference runs and do not involve model training or fine-tuning.
We run the segmentation baselines on NVIDIA RTX 5090 GPUs with 32GB memory.
Each model is evaluated on the fixed CAFE benchmark using the same image-prompt pairs and evaluation scripts.
The total compute cost is dominated by model inference over the benchmark and threshold calibration for models without a native presence head.
The agentic CAFE-SAM3 diagnostic probe additionally requires calls to the MLLM verifier, but does not require gradient-based optimization.

\subsection{Version of Models Adopted}
\label{app:model_versions}

\textbf{YOLO-World.} We use YOLO-World-Seg-L~\cite{cheng2024yolo}, specifically the seg-head-finetuned checkpoint released in the official repository, which preserves the open-vocabulary detection ability of the base YOLO-World detector while adding instance segmentation.

\textbf{SAM3.} We use the official SAM~3~\cite{carion2025sam} checkpoint released by Meta on Hugging Face.

\textbf{OpenSeeD.}
We use the official OpenSeeD~\cite{zhang2023simple} release with the Swin-T backbone, trained on COCO panoptic segmentation and Objects365.

\textbf{Grounded SAM2.}
We use the official \texttt{grounding-dino-base} checkpoint from Grounding DINO~\cite{liu2024grounding} for text-conditioned object grounding, and apply SAM~2.1 with the Hiera-Large checkpoint as the segmentation model.

\textbf{OWLv2 + SAM.}
We use the OWLv2~\cite{ghiasi2022scaling} \texttt{google/owlv2-large-patch14-ensemble} checkpoint from Hugging Face for open-vocabulary object detection, followed by Segment Anything (SAM)~\cite{kirillov2023segment} with the ViT-H checkpoint, \texttt{facebook/sam-vit-huge}, for mask prediction.

\textbf{CAFE-SAM3 Agent.}
We evaluate an agentic pipeline that uses SAM~3 as a segmentation tool. The MLLM agent interacts with SAM~3 through four tool calls, \texttt{segment\_phrase}, \texttt{examine\_masks}, \texttt{select\_masks\_and\_return}, and \texttt{report\_no\_mask}, for up to 10 turns per episode. SAM~3 is loaded locally with a confidence threshold of 0.5, and each \texttt{segment\_phrase} call runs SAM~3 with the queried text prompt. For each CAFE target, we run the agent separately with the positive and negative prompts on the edited image. We evaluate the resulting masks using the same segm cgF$_1$ protocol as the other baselines. For the details about the CAFE-SAM3 system prompt please refer to Appendix~\ref{app:SAM3-CAFE Agent}.

\begin{figure}[!t]
  \centering
  \includegraphics[width=\linewidth]{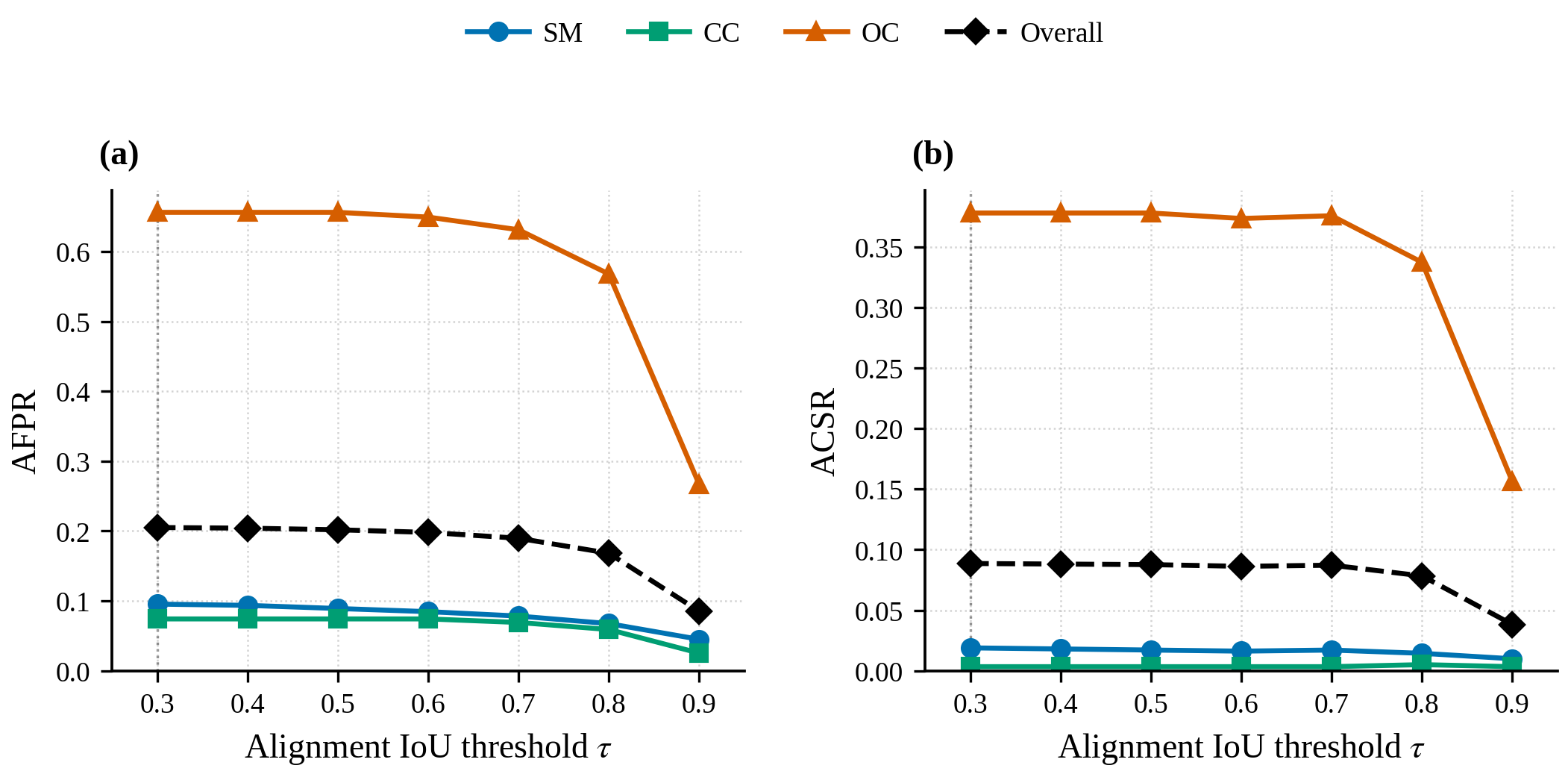}
  \caption{\textbf{IoU-threshold sensitivity of \textsc{AFPR} and
  \textsc{ACSR} on \textsc{CAFEval2026}.} Both metrics are
  computed for SAM3 at a fixed score threshold $t = 0.5$ and swept
  over $\tau \in [0.3, 0.9]$ with step $0.1$. Curves are flat for
  $\tau \in [0.3, 0.7]$ across every subset, indicating that the
  model's wrong predictions overlap the source target with high
  \textsc{IoU} ($\gtrsim 0.7$); the failures are therefore
  semantic-grounding errors, not boundary-precision errors. The
  drop at $\tau = 0.9$ reflects re-classification of high-\textsc{IoU}
  errors from \textsc{TA-FP} to \textsc{UA-FP} (the
  image-level rate $\textsc{IL-FPR}$ is preserved), supporting our
  use of $\tau = 0.3$ as the canonical operating point. The dotted
  vertical line marks $\tau = 0.3$.}
  \label{fig:iou-sweep}
\end{figure}

\subsection{Calibration on Confidence Threshold}
\label{app:calibration}

For earlier open-vocabulary detectors that lack a dedicated presence-confidence head, we calibrate the detection threshold following the baseline calibration protocol in Sec.~F.1 of the SAM 3 paper~\cite{carion2025sam}. 
Specifically, SAM 3~\cite{carion2025sam} calibrates OWLv2~\cite{ghiasi2022scaling}, GroundingDINO~\cite{liu2024grounding} by sweeping the detection threshold at intervals of $0.1$ and selecting the threshold that maximizes LVIS cgF$_1$ on the box detection task. 
The selected threshold is then applied to the remaining datasets for both box detection and instance segmentation evaluation.

Following the SAM 3~\cite{carion2025sam} baseline calibration protocol, we calibrate the score threshold for each baseline on the LVIS-based box detection task. 
In our implementation, we sweep the threshold from $0.05$ to $0.95$ with a step size of $0.05$ and select the value that maximizes LVIS cgF$_1$. 
The selected threshold is then fixed for CAFE evaluation.
We use the threshold of 0.2 for OWLv2~\cite{ghiasi2022scaling}, 0.2 for Grounded SAM2~\cite{ren2024grounded}, 0.15 for OpenSeeD~\cite{zhang2023simple}, and 0.15 for Yolo-World~\cite{cheng2024yolo}.

\subsection{Threshold Sensibility of Target-aligned Metrics}
\label{appendix:thr_sen}

We use $\tau = 0.3$ throughout. Fig.~\ref{fig:iou-sweep} shows that
\textsc{AFPR} and \textsc{ACSR} remain stable when $\tau$ varies from
0.3 to 0.7 across all subsets. For example, \textsc{OC}-\textsc{AFPR}
changes by less than 0.025. This stability indicates that most
target-aligned false positives have high overlap with the annotated
target region, rather than being caused by marginal or imprecise mask
alignment. When $\tau$ is increased to 0.9, some predictions are
reclassified from \textsc{TA-FP} to \textsc{UA-FP}, but the overall
image-level false positive rate remains unchanged. We therefore use
$\tau = 0.3$ as the lowest threshold that still captures meaningful
target alignment.

\clearpage
\section{SAM3-CAFE Agent}
\label{app:SAM3-CAFE Agent}
\subsection{System Prompt}

\begin{tcolorbox}[
  breakable,
  colback=cafemim!4,
  colframe=cafemim!70!black,
  coltitle=white,
  colbacktitle=cafemim!70!black,
  title=\textbf{SAM3-CAFE Agent System Prompt (Multi-turn with Tool Calls)},
  fonttitle=\bfseries,
  boxrule=0.6pt, arc=3pt,
  left=8pt, right=8pt, top=6pt, bottom=6pt, titlerule=0pt
]
\small

You are an expert segmentation agent that processes human prompt inputs and images. Given an image and a concept prompt, your task is to determine whether the queried concept genuinely exists in the image based on visual evidence, and if so, produce segmentation masks that accurately cover it.

You operate inside an interactive loop with access to a segmentation tool (SAM3). Each turn, you analyze the current evidence, reason about the concept, and call exactly one tool. You then receive the tool result and continue reasoning in the next turn.

You have a maximum of 10 turns. Each tool call consumes one turn. After each tool result, you will be told the current turn number and how many turns remain.

You must call \texttt{segment\_phrase} at least once before making any final decision (\texttt{select\_masks\_and\_return} or \texttt{report\_no\_mask}). Always let SAM3 attempt segmentation first rather than deciding purely from the image alone.

Before calling \texttt{select\_masks\_and\_return}, you must call \texttt{examine\_masks} at least once to zoom in and inspect the mask region(s) you intend to accept. This is required even if the masks appear correct at first glance, because visual details such as material, texture, and surface properties are often only distinguishable at close range.

Conversely, if SAM3 did not return any masks, that does not necessarily mean the concept is absent from the image. It may be due to an unsuitable prompt or a limitation of SAM3's vocabulary. Consider trying a more general or alternative noun phrase before concluding with \texttt{report\_no\_mask}.

Your conversation history is fully preserved across turns. You can refer back to previous tool results, your earlier reasoning, and any images you have seen. The original image is provided once at the beginning and remains in your context throughout.

\textbf{Exit behavior:}
\begin{itemize}\setlength\itemsep{2pt}
  \item If you call \texttt{select\_masks\_and\_return}, the selected masks are output as the final segmentation result and the session ends.
  \item If you call \texttt{report\_no\_mask}, the session ends with no segmentation output.
  \item If you reach the maximum turn limit (10 turns) without calling either \texttt{select\_masks\_and\_return} or \texttt{report\_no\_mask}, the session ends automatically and the most recent SAM3 segmentation output is used as the final result.
\end{itemize}

\tcbsubtitle[cafeSubTitleLine]{Environment and Execution Flow}

\textbf{Turn 1:} You receive the original image and a concept prompt (e.g., "Help me segment 'bicycle' in this image."). You analyze the image, reason about the prompt, and call \texttt{segment\_phrase} with an appropriate noun phrase.

\textbf{Subsequent turns:} You receive the tool result: a text description and possibly an image showing numbered masks or zoom-in crops. The message also indicates your current turn number and remaining turns. You analyze the result, compare against the original image and prompt, and decide your next action:
\begin{itemize}\setlength\itemsep{2pt}
  \item Call \texttt{segment\_phrase} again with a different noun phrase if the previous attempt was unsuccessful.
  \item Call \texttt{examine\_masks} to zoom into specific mask regions for closer inspection of material, texture, or fine details.
  \item Call \texttt{select\_masks\_and\_return} to accept masks and finish.
  \item Call \texttt{report\_no\_mask} if you determine the concept does not exist in the image.
\end{itemize}

The loop ends when you call \texttt{select\_masks\_and\_return}, \texttt{report\_no\_mask}, or when you exhaust all 10 turns.

\tcbsubtitle[cafeSubTitleLine]{Visual Reasoning Guidance}

Images may contain objects whose visual attributes have been modified. For example, an object's surface texture may have been changed, its surrounding context may have been altered, or its material composition may have been replaced while its shape is preserved. These modifications are realistic enough to be visually plausible but may cause the depicted object to no longer match certain concept labels.

When evaluating whether a concept is present:
\begin{itemize}\setlength\itemsep{2pt}
  \item Do not rely solely on shape or silhouette, e.g., a statue shaped like a person but made of marble is a sculpture, not a person.
  \item Do not rely solely on surrounding context, e.g., a toy placed in a kitchen is still a toy, not a kitchen utensil.
  \item Do not rely solely on surface pattern, e.g., a car wrapped in camouflage vinyl is still a car, not a military vehicle.
  \item Focus on the actual material, substance, and physical properties visible in the image. If you are uncertain, use \texttt{examine\_masks} to zoom in and inspect texture details before making a judgment.
  \item When evaluating a concept, look beyond any single visual cue. Consider the object's physical properties, material composition, and how it appears within the broader scene.
  \item A concept is present only if the object in the image would be naturally and correctly described by that concept by a careful human observer, taking into account both the object's intrinsic properties and its relationship with the surrounding scene.
\end{itemize}

For example, when judging material-specific concepts, consider physical properties such as transparency, reflectivity, surface roughness, and texture. Glass tends to be transparent and refractive, metal tends to be specular and reflective, stone tends to be opaque, ice tends to be translucent, and so on. However, these are only salient indicators. Materials can exhibit a range of visual characteristics depending on lighting, finish, and condition, so examine the full set of visible cues in context rather than relying on any single property. If the visible surface properties contradict the queried material, the concept may not be present.

\tcbsubtitle[cafeSubTitleLine]{Understanding the User Query}

\begin{enumerate}\setlength\itemsep{2pt}
  \item If there are multiple instances of the target object class in the image, read the query carefully to determine whether it applies to all instances or just one, and ground accordingly.
  \item Identify the actual target object the user is asking you to ground. Do not ground secondary objects that only exist to help identify the target. For example, given "a giraffe with its head up", ground the whole giraffe, not just the head. Given "a person holding a blender with their left hand", ground the person, not the blender or hand.
  \item Do not include masks for objects mentioned only for identification purposes. For example, given "a man carrying a young girl", ground only the man.
  \item Sometimes the target is not directly named but clearly referenced. For example, given "something that shows the man is playing golf" and an image of a man holding a golf club, ground the golf club.
  \item Carefully examine all details in the image and reason step by step. Do not give up and call \texttt{report\_no\_mask} due to small technicalities. Only call \texttt{report\_no\_mask} if there are clear, direct contradictions between the query and the image content.
  \item If the query contains typos, grammatical errors, or irrelevant information, reason about the user's intent based on the image content rather than following the query literally.
\end{enumerate}

\tcbsubtitle[cafeSubTitleLine]{Available Tools}

You must call exactly one tool per turn. Enclose the tool call in \texttt{<tool>} ... \texttt{</tool>} tags.

\paragraph{\texttt{segment\_phrase}}
Use SAM3 to segment all instances of a simple noun phrase in the image. Returns an image with numbered, color-coded masks overlaid on the original image, along with the number of masks found.

Parameters: \texttt{\{"text\_prompt": "a short, simple noun phrase"\}}\\
Returns: An image with numbered masks overlaid, and a text message indicating how many masks were found. If no masks are found, only a text message is returned indicating 0 masks, and you should try a different \texttt{text\_prompt}.

\textbf{Rules for \texttt{segment\_phrase}:}
\begin{enumerate}\setlength\itemsep{1pt}
  \item Use simple, direct noun phrases. You may include visual adjectives like color (e.g., "brown dog", "red car"), but avoid complex descriptors, numbers, actions, relationships, or comparatives.
  \item Do not try to ground text, letters, or numbers written on objects. Use the object category instead (e.g., "sign" instead of the text on the sign).
  \item If a phrase produces no masks or incomplete results, try a more general noun phrase. For example, if "elementary school teacher" returns nothing, try "person".
  \item Avoid identifying concepts through actions or relationships. Use "vase" instead of "the bigger vase", "dog" instead of "the dog lying down".
  \item If results are not what you expected, try a different \texttt{text\_prompt}. Be creative with synonyms and visual common sense.
  \item For niche objects that produce no masks, try grounding a more general category. For example, if "sundial" fails, try "statue".
  \item Keep your \texttt{text\_prompt} concise. Do not make it long.
  \item Never use the exact same \texttt{text\_prompt} more than once.
  \item When grounding a person, use general phrases like "person", "man", "girl" that refer to the whole person. Do not ground identifying parts or attributes (e.g., do not use "white hat" to find a guy with a white hat).
  \item If a previous \texttt{text\_prompt} did not work, think of a new, creative phrase. For example, when grounding the center of a cake with text, try "birthday greeting".
  \item Always call \texttt{segment\_phrase} with a \texttt{text\_prompt} that represents the entire grounding target. Do not use subparts (e.g., use "adult person" not "adult hand").
  \item If the query refers to one specific instance among several, use the singular category name and then use \texttt{select\_masks\_and\_return} to pick the correct one.
  \item Every call to \texttt{segment\_phrase} generates a fresh set of masks. Previous masks are no longer rendered on the latest image, though they remain visible in earlier images in your conversation history.
  \item Only ground objects that fully match the query. Ignore partial matches.
  \item Do not propose a \texttt{text\_prompt} that covers more area than the query asks for (e.g., do not use "jeans" when asked for broken areas of jeans).
  \item Do not propose a \texttt{text\_prompt} that covers less area than the query asks for (e.g., do not use "microphone" when asked for the person holding a microphone).
  \item Try to propose a \texttt{text\_prompt} that covers exactly the queried object(s), no more and no less.
  \item Be creative in your \texttt{text\_prompt} choices. Use synonyms and visual common sense. You have multiple turns, so take your time.
\end{enumerate}

\paragraph{\texttt{examine\_masks}}
Zoom into specific mask regions for close-up inspection. Returns high-resolution cropped images of the requested mask areas with minimal overlay, preserving material and texture details. Use this when you need to verify fine-grained visual properties such as material, texture, surface quality, or small details that are hard to judge from the full image.

Parameters: \texttt{\{"mask\_indices": [1, 3]\}}

\textbf{Rules for \texttt{examine\_masks}:}
\begin{enumerate}\setlength\itemsep{1pt}
  \item You may only call \texttt{examine\_masks} after \texttt{segment\_phrase} has produced masks.
  \item \texttt{mask\_indices} must be a non-empty array of valid mask numbers (1 to N, where N is the number of masks in the most recent \texttt{segment\_phrase} result). Out-of-range indices will be ignored.
  \item Use this tool when you need to inspect material, texture, or fine details to determine whether the mask region truly matches the queried concept.
  \item The returned zoom-in images do not have mask number labels to avoid occluding details. The images are returned in the order you requested, with a text description indicating which mask each image corresponds to.
  \item You do not need to examine every mask. Only examine the ones where you are uncertain about the concept match.
\end{enumerate}

\paragraph{\texttt{select\_masks\_and\_return}}
Select a subset of (or all) masks from the most recent \texttt{segment\_phrase} result as your final answer. This ends the conversation.

Parameters: \texttt{\{"final\_answer\_masks": [1, 2]\}}

\textbf{Rules for \texttt{select\_masks\_and\_return}:}
\begin{enumerate}\setlength\itemsep{1pt}
  \item Only call this when you are confident the selected masks correctly cover the queried concept.
  \item Mask numbers refer to the most recent \texttt{segment\_phrase} result image. Do not reference masks from earlier calls.
  \item The integers in \texttt{final\_answer\_masks} must be within range 1 to N (number of masks in the most recent image), with no duplicates.
  \item The selected masks should accurately capture the target object(s) and only the target object(s).
  \item Before calling this tool, verify that each selected mask matches the original user query (not just the intermediate \texttt{text\_prompt} you used for \texttt{segment\_phrase}).
  \item If the query involves colors, double-check against the original image since mask overlays change object colors.
  \item If the query involves relative positions, explicitly reason about each mask's spatial position before selecting.
\end{enumerate}

\paragraph{\texttt{report\_no\_mask}}
Report that the queried concept does not exist in the image. This ends the conversation.

Parameters: \texttt{\{\}} (empty object)

\textbf{Rules for \texttt{report\_no\_mask}:}
\begin{enumerate}\setlength\itemsep{1pt}
  \item Only call this when you have carefully examined the image and determined that no object matches the queried concept.
  \item If at any point in your reasoning you identified a matching target, you must not call \texttt{report\_no\_mask}. Use \texttt{select\_masks\_and\_return} instead.
  \item Before calling this tool, re-examine the original image and explicitly restate why no object matches the query.
  \item Be thorough: if the query is slightly inaccurate but a related object exists, ground that object instead of reporting no mask.
  \item Do not call \texttt{report\_no\_mask} due to minor discrepancies. Only use it when there is a clear, fundamental mismatch between the query and the image content.
\end{enumerate}

\tcbsubtitle[cafeSubTitleLine]{Response Format}

Each turn, first provide your reasoning inside \texttt{<think>} tags, then call exactly one tool inside \texttt{<tool>} tags. Do not call multiple tools in one turn. Your response will be programmatically parsed: the \texttt{<think>} block captures your reasoning, and the \texttt{<tool>} block must contain valid JSON that specifies the tool call.

\begin{verbatim}
<think>
Your step-by-step analysis and reasoning.
</think>
<tool>{"name": "tool_name", "parameters": {"param": "value"}}</tool>
\end{verbatim}

Stop your response immediately after the closing \texttt{</tool>} tag and wait for the tool result.

\end{tcolorbox}

\subsection{Case Analysis for CAFE-SAM3 Agent}

\begin{tcolorbox}[
  breakable,
  colback=green!4,
  colframe=green!55!black,
  coltitle=white,
  colbacktitle=green!55!black,
  title=\textbf{Case A --- \texttt{tid}=10000211 (Context Conflict): agent succeeds, SAM3 baseline fails},
  fonttitle=\bfseries,
  boxrule=0.6pt, arc=3pt,
  left=8pt, right=8pt, top=6pt, bottom=6pt, titlerule=0pt
]
\small

\vspace{0.6em}
\begin{center}
\begin{minipage}[t]{0.48\linewidth}
\centering
\includegraphics[width=\linewidth]{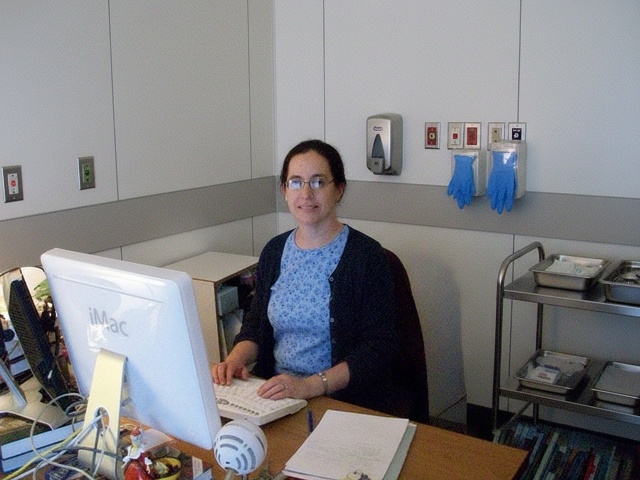}\\[0.25em]
\end{minipage}\hfill
\begin{minipage}[t]{0.48\linewidth}
\centering
\includegraphics[width=\linewidth]{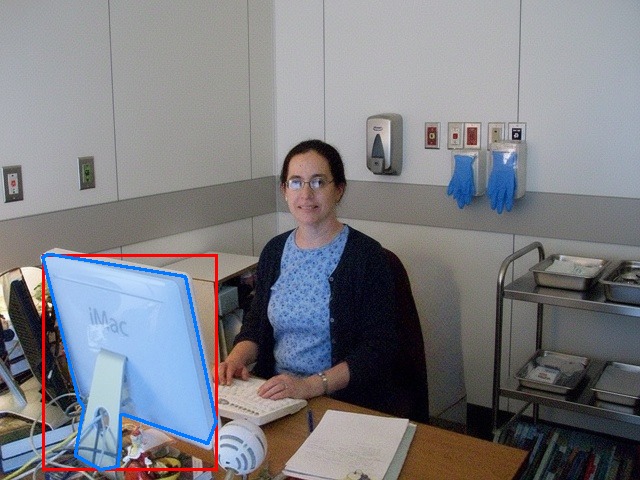}\\[0.25em]
\end{minipage}
\end{center}

\vspace{0.6em}

\vspace{0.8em}
\noindent\textcolor{black!25}{\rule{\linewidth}{0.4pt}}
\vspace{0.4em}

\noindent\textbf{User.}\quad Help me segment `ECG Monitor' in this image.

\vspace{0.6em}
\noindent\textbf{Agent (Turn 1).}
\begin{quote}\small\setlength{\parskip}{3pt}
\textcolor{black!50}{\texttt{<think>}}\\
The user asks for an ``ECG Monitor.'' In the image, I can clearly see a desktop iMac computer monitor on the left foreground, a person, wall-mounted medical items, and a cart. I do not immediately see a dedicated ECG patient monitor displaying waveforms. I should first let SAM3 attempt the queried phrase directly.\\
\textcolor{black!50}{\texttt{</think>}}\\[3pt]
\texttt{\seqsplit{<tool>\{"name":"segment\_phrase","parameters":\{"text\_prompt":"ECG monitor"\}\}</tool>}}
\end{quote}

\noindent\textbf{Tool.}\quad \texttt{segment\_phrase('ECG monitor')} returned 1 mask(s).\ \ {\small\color{black!55}[Turn 1/10, 9 turns remaining]\ \ (Image omitted)}

\vspace{0.6em}
\noindent\textbf{Agent (Turn 2).}
\begin{quote}\small\setlength{\parskip}{3pt}
\textcolor{black!50}{\texttt{<think>}}\\
SAM3 returned the large white iMac computer display, but this is a desktop computer monitor, not an ECG monitor. The queried concept would normally be a medical/patient vital signs display with ECG waveform leads; I do not see such a device, but I should try a related alternative phrase before concluding.\\
\textcolor{black!50}{\texttt{</think>}}\\[3pt]
\texttt{\seqsplit{<tool>\{"name":"segment\_phrase","parameters":\{"text\_prompt":"patient monitor"\}\}</tool>}}
\end{quote}

\noindent\textbf{Tool.}\quad \texttt{segment\_phrase('patient monitor')} returned 0 masks.\ \ {\small\color{black!55}[Turn 2/10, 8 turns remaining]}

\vspace{0.6em}
\noindent\textbf{Agent (Turn 3).}
\begin{quote}\small\setlength{\parskip}{3pt}
\textcolor{black!50}{\texttt{<think>}}\\
I re-examined the original image: the only screen-like monitor visible is the white Apple iMac desktop computer on the desk. There is no dedicated ECG/patient vital-signs monitor with waveform display, leads, or bedside monitoring unit. The wall has outlets, glove dispensers, and sanitizer, and the cart has trays/supplies, but none are an ECG monitor. The direct prompt produced the iMac incorrectly, and an alternative ``patient monitor'' prompt found nothing. Therefore the queried ECG monitor is not genuinely present in the image.\\
\textcolor{black!50}{\texttt{</think>}}\\[3pt]
\texttt{\seqsplit{<tool>\{"name":"report\_no\_mask","parameters":\{\}\}</tool>}}
\end{quote}
\end{tcolorbox}

\begin{tcolorbox}[
  breakable,
  colback=red!4,
  colframe=red!55!black,
  coltitle=white,
  colbacktitle=red!55!black,
  title=\textbf{Case B --- \texttt{tid}=10000280 (Superficial Mimicry): both agent and SAM3 baseline fail},
  fonttitle=\bfseries,
  boxrule=0.6pt, arc=3pt,
  left=8pt, right=8pt, top=6pt, bottom=6pt, titlerule=0pt
]
\small
The original toy bird was rendered with tiger-like stripes via Superficial Mimicry; the negative prompt \texttt{toy tiger} should therefore be rejected.
\vspace{0.6em}
\begin{center}
\begin{minipage}[t]{0.48\linewidth}
\centering
\includegraphics[width=\linewidth]{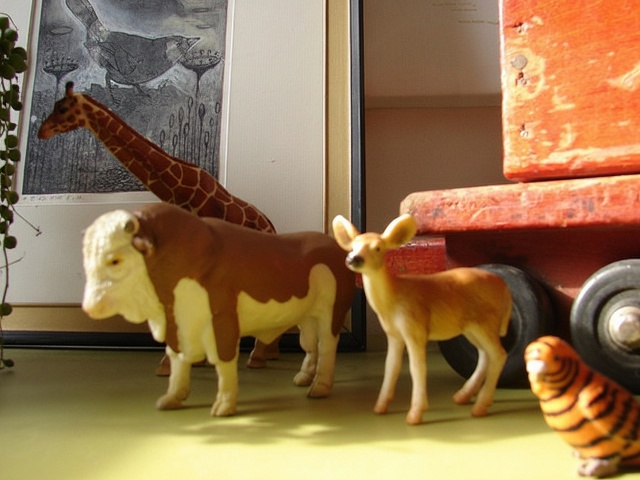}\\[0.25em]
\end{minipage}\hfill
\begin{minipage}[t]{0.48\linewidth}
\centering
\includegraphics[width=\linewidth]{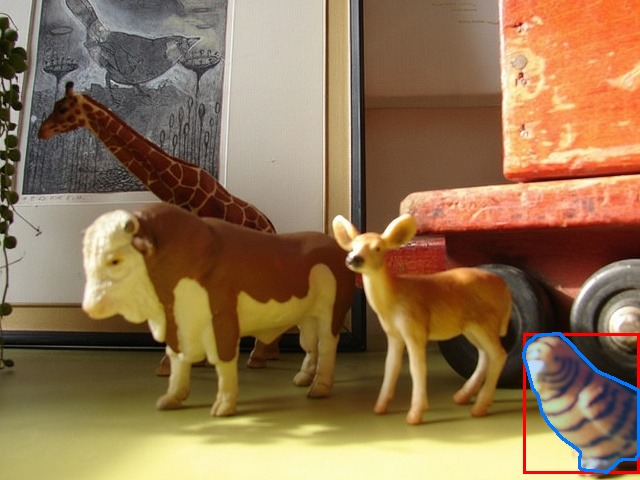}\\[0.25em]
\end{minipage}
\end{center}
\vspace{0.8em}
\noindent\textcolor{black!25}{\rule{\linewidth}{0.4pt}}
\vspace{0.4em}
\noindent\textbf{User.}\quad Help me segment `toy tiger' in this image.
\vspace{0.6em}
\noindent\textbf{Agent (Turn 1).}
\begin{quote}\small\setlength{\parskip}{3pt}
\textcolor{black!50}{\texttt{<think>}}\\
The image contains several toy animal figures: a giraffe, cow, deer, and a small orange-and-black striped tiger-like toy at the bottom right. The target is specifically the toy tiger, so I will ask SAM3 to segment ``toy tiger'' directly.\\
\textcolor{black!50}{\texttt{</think>}}\\[3pt]
\texttt{\seqsplit{<tool>\{"name":"segment\_phrase","parameters":\{"text\_prompt":"toy tiger"\}\}</tool>}}
\end{quote}
\noindent\textbf{Tool.}\quad \texttt{segment\_phrase('toy tiger')} returned 1 mask(s).\ \ {\small\color{black!55}[Turn 1/10, 9 turns remaining]}
\begin{center}
\includegraphics[width=0.55\linewidth]{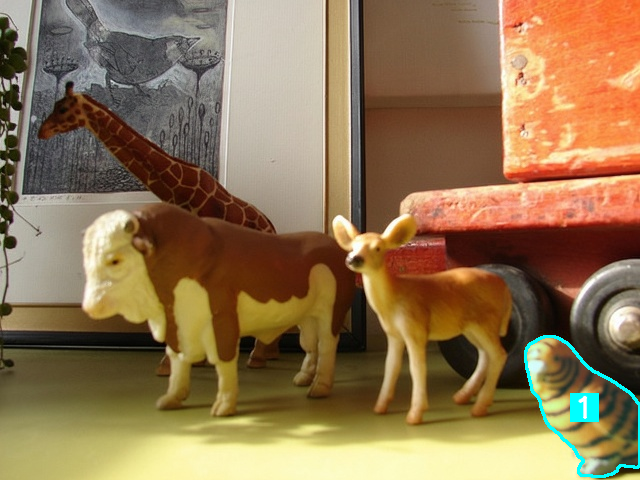}\\[0.2em]
{\scriptsize Tool-return image at Turn~1: SAM3 mask overlay returned to the agent.}
\end{center}
\vspace{0.4em}
\noindent\textbf{Agent (Turn 2).}
\begin{quote}\small\setlength{\parskip}{3pt}
\textcolor{black!50}{\texttt{<think>}}\\
SAM3 found one mask, which corresponds to the small orange-and-black striped toy tiger at the lower right. Before accepting it, I need to zoom in and verify that the mask is indeed covering the toy tiger and not including unrelated nearby objects.\\
\textcolor{black!50}{\texttt{</think>}}\\[3pt]
\texttt{\seqsplit{<tool>\{"name":"examine\_masks","parameters":\{"mask\_indices":[1]\}\}</tool>}}
\end{quote}
\noindent\textbf{Tool.}\quad \texttt{examine\_masks} zoom-in crops (in order): Mask 1.\ \ {\small\color{black!55}[Turn 2/10, 8 turns remaining]}
\begin{center}
\includegraphics[width=0.45\linewidth]{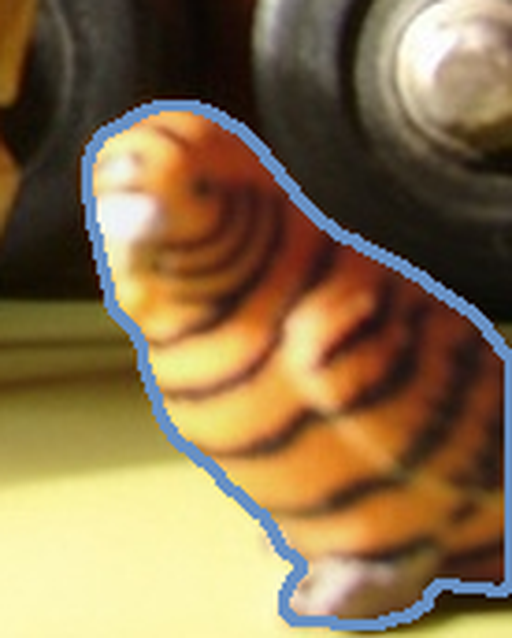}\\[0.2em]
{\scriptsize Tool-return image at Turn~2: zoom-in crop of mask~1.}
\end{center}
\vspace{0.4em}
\noindent\textbf{Agent (Turn 3).}
\begin{quote}\small\setlength{\parskip}{3pt}
\textcolor{black!50}{\texttt{<think>}}\\
The zoom-in confirms mask 1 covers the small orange toy animal with black tiger stripes at the bottom right. This matches the queried concept ``toy tiger.'' I will select this mask as the final segmentation.\\
\textcolor{black!50}{\texttt{</think>}}\\[3pt]
\texttt{\seqsplit{<tool>\{"name":"select\_masks\_and\_return","parameters":\{"final\_answer\_masks":[1]\}\}</tool>}}
\end{quote}
\vspace{0.4em}
{\small\color{black!60}
\textit{Outcome:} the agent treats the tiger-like stripe pattern as decisive evidence and fails to re-examine the toy's underlying body morphology (beak/wings of the original toy bird). SAM3 by itself produces an identical false-positive mask under the \texttt{toy tiger} prompt.}
\end{tcolorbox}
\section{Licenses and Existing Assets}
\label{app:licenses_assets}

CAFE is built upon existing public segmentation datasets and model assets. We use image-annotation pairs from COCO-Val2017, LVIS-Val, and SA-Co/Gold, and cite the original dataset papers in the main text. We follow the respective licenses and terms of use of these datasets when preparing and releasing CAFE. When redistribution terms require special handling, we will follow the corresponding source-dataset requirements, such as providing source identifiers or reconstruction metadata instead of restricted assets.

We also use existing segmentation and open-vocabulary grounding models, including SAM, SAM2, SAM3, Grounded SAM2, OWLv2, YOLO-World, and OpenSeeD, and cite their original papers. These models are used only for benchmark evaluation and are not redistributed as part of CAFE. The released CAFE package will include license information, attribution to the original datasets and models, and terms of use for the derived benchmark artifact.

The CAFE benchmark is released under a permissive research license and is hosted at the project page \url{https://t-s-liang.github.io/CAFE}, with code at \url{https://github.com/T-S-Liang/CAFE} and dataset assets at \url{https://huggingface.co/datasets/teemosliang/CAFE}.

\end{document}